\documentclass[3p,times]{elsarticle}

\usepackage{array}
\usepackage{fontawesome5}
\usepackage{ccicons}
\usepackage{pifont}
\usepackage{booktabs}
\usepackage{longtable,tabularx,ltxtable,ragged2e}
\usepackage{xltabular}
\usepackage{cellspace}
\usepackage{tabularray}
\newcommand\cincludegraphics[2][]{\raisebox{-0.2\height}{\includegraphics[#1]{#2}}}

\newcolumntype{Y}{>{\RaggedRight\arraybackslash}X}
\newcolumntype{P}[1]{>{\centering\arraybackslash}p{#1}}

\usepackage{filecontents}

\usepackage{adjustbox}
\usepackage{graphicx}
\usepackage{flushend} 
\usepackage{ccicons}
\usepackage{mwe}
\usepackage{makerobust} 
\MakeRobustCommand\rotatebox 
\graphicspath{{./pdf/}{./jpeg/}{./images/}}
\DeclareGraphicsExtensions{.pdf,.jpeg,.jpg,.png}
\DeclareGraphicsExtensions{.pdf,.jpeg,.jpg,.png} 
\usepackage{amssymb}
\usepackage{pifont}
\newcommand{\cmark}{\ding{51}}%

\usepackage{pseudocode}
\usepackage{verbatim}
\usepackage{sverb}

\usepackage{amsmath}
\usepackage[ruled]{algorithm2e}
\usepackage[noend]{algpseudocode}

\usepackage{caption} 
\usepackage{subcaption}
\usepackage[nodots,nocompress]{numcompress}

\usepackage{textcomp}

\usepackage{upgreek}

\usepackage{url}
\usepackage{float}

\usepackage{amsmath}

\usepackage[T1]{fontenc}

\usepackage{array}
\usepackage{booktabs}
\usepackage{color}
\usepackage{xcolor}

\usepackage{units}

\usepackage{soul}
\usepackage{color}
	\definecolor{celadon}{rgb}{0.67, 0.88, 0.69}
  \definecolor{flamingopink}{rgb}{0.99, 0.56, 0.67}

\usepackage{makecell}

\usepackage{enumitem}

\usepackage{listings}

\biboptions{numbers}

\begin{document}

\begin{frontmatter}


\title{Computer Vision for Multimedia Geolocation in Human\\Trafficking Investigation: A Systematic Literature Review}

\author[add1]{Opeyemi Bamigbade}
\ead{opeyemi.bamigbade@postgrad.wit.ie}
\author[add2]{Mark Scanlon}
\ead{mark.scanlon@ucd.ie}
\author[add1]{John Sheppard}
\ead{John.Sheppard@setu.ie}
\address[add1]{South East Technological University, Waterford, Ireland}
\address[add2]{Forensics and Security Research Group, School of Computer Science, University College Dublin, Ireland}


\begin{abstract}
The task of multimedia geolocation is becoming an increasingly essential component of the digital forensics toolkit to effectively combat human trafficking, child sexual exploitation, and other illegal acts. Typically, metadata-based geolocation information is stripped when multimedia content is shared via instant messaging and social media. The intricacy of geolocating, geotagging, or finding geographical clues in this content is often overly burdensome for investigators. Recent research has shown that contemporary advancements in artificial intelligence, specifically computer vision and deep learning, show significant promise towards expediting the multimedia geolocation task. This systematic literature review thoroughly examines the state-of-the-art leveraging computer vision techniques for multimedia geolocation and assesses their potential to expedite human trafficking investigation. This includes a comprehensive overview of the application of computer vision-based approaches to multimedia geolocation, identifies their applicability in combating human trafficking, and highlights the potential implications of enhanced multimedia geolocation for prosecuting human trafficking. 123 articles inform this systematic literature review. The findings suggest numerous potential paths for future impactful research on the subject.
\end{abstract}

\end{frontmatter}

\section{Introduction}
In this digital age where information flows across borders with unprecedented ease, the fight against human trafficking has encountered a formidable adversary. Human traffickers exploit the very tools that connect our world to perpetrate their heinous crimes, leaving victims in the shadows of an interconnected society. The Global Estimates of Modern Slavery Report highlighted that human trafficking affects 25 million people and makes more than \$150 billion US dollars in illicit profits annually worldwide and is showing an increasing trend~\cite{dimas2022_Operations-research, Kshetri_regulatory_2023}. As the battle to combat this global scourge intensifies, artificial intelligence shows great promise to aid law enforcement in the expedition of digital forensic investigations, particularly in multimedia geolocation, which has grown greatly in the recent decade as a result of the rapid expansion of contemporary machine learning capabilities~\cite{Glistrup_Urban-Image_2022, Yokota_Visual-based_2020, aronson_computer_2018}.

Multimedia geolocation is determining the real-world location in which a multimedia file, such as an image or video, has been captured. While this task can be straightforward should metadata/EXIF data be present, it is often difficult, or impossible, to achieve. This is due to many multimedia found on the internet and/or at crime scenes have these metadata tags stripped. Automatic, semantic multimedia geolocation has the potential to open up new possibilities across a range of lawful investigations and across a variety of other fields. Knowing the location of multimedia data facilitates the easy determination of hundreds of additional attributes~\cite{Nam_Revisiting_2017}, such as population density estimation or land cover estimation~\cite{Choi_Multimodal_2015}. It has the potential to transform law enforcement, particularly because both partial and full automation would be possible~\cite{Friedland_Multimodal_2010, Ivanov_Comparative-Study_2013, aronson_computer_2018}.

Artificial intelligence is one viable solution to alleviate the digital forensics backlog that has become commonplace in law enforcement digital forensic laboratories worldwide~\cite{moreira_image_2018, scanlon2023ChatGPTforDigitalForensics}. A growing amount of digital research, in collaboration with advances in computer vision, and machine learning, has enabled progressive techniques to identify and process information contained in multimedia evidence~\cite{Zerdoumi_A-new-spatial_2022} which means that we can now acquire more specific insights about the context of an image and the interactions between items inside it, allowing us to better handle these difficulties~\cite{Keerthi_RIPA-2018}. With such a breakthrough, law enforcement efforts and combat towards the interception of victims in transition, the reduction of human trafficking activities, and increasing prosecution of such criminals will be more pronounced.

This systematic review focuses on the critical role of computer vision techniques in multimedia geolocation applied to human trafficking investigations, highlighting other notable methodologies and approaches to extracting geolocation information from text, images, and videos that could be used to review hidden networks of traffickers, locating victims and bringing these criminals to justice. It navigates the digital evidence analysis and the effect of variational factors in recognition algorithms, which is a major drawback to effective geolocation. The primary contributions of this research are as follows: 
\begin{itemize}
\item A thorough and systematic review of the extant literature on State-of-the-Art multimedia geolocation techniques, 
\item Identification of multimedia evidence enhancements with computer vision approaches from a digital forensics perspective, 
\item Accentuate the implications of multimedia geolocation for evidence collection and prosecution in human trafficking cases, 
\item Highlights of major drawbacks in the advancement of geolocation estimation and future directions.
\end{itemize}

A systematic literature review that provides an objective approach to identifying the nature and extent of research on this topic has yet to be carried out to date. To the best of our knowledge, this is the first systematic review study focusing on how computer vision is used in multimedia geolocation applied to human trafficking investigations. We hope that by shedding light on the intersection of technology and human rights, we may contribute to the ongoing worldwide struggle for multimedia content geolocation estimation, reduce human trafficking activities, and bring hope to those who are victims of this modern-day scourge.

This paper is organized as follows. Section~\ref{sec:methodology} describes the research methodology, including the research questions and themes categorization. Section~\ref{sec:HT} provides some context around the prevalence of human trafficking. Section~\ref{sec:backgroundCV} provides the background of computer vision and its constituent techniques. Section~\ref{sota_multimedia_geolocation} focuses on the state of the art in multimedia geolocation and Section~\ref{sec:usecases} highlights multimedia geolocation use cases. Section~\ref{sec:forensicsHT} presents the state of multimedia forensics in human trafficking investigations. Section~\ref{sec:discussion} discusses the main findings of the study. Lastly, Section~\ref{sec:conclusion} outlines the conclusions and future directions.

\section{Methodology for Systematic Literature Review}
\label{sec:methodology}
\subsection{Defining Research Questions} 
The purpose of these research questions is to guide our investigation into the current state of knowledge, identify research gaps, and provide valuable insights for future developments in the field of computer vision-based multimedia geolocation in the context of digital forensics. They also serve as the basis for conducting an extensive and systematic literature review.
\begin{figure*}[h!]
\centering
\includegraphics[width=17cm, height=12cm]{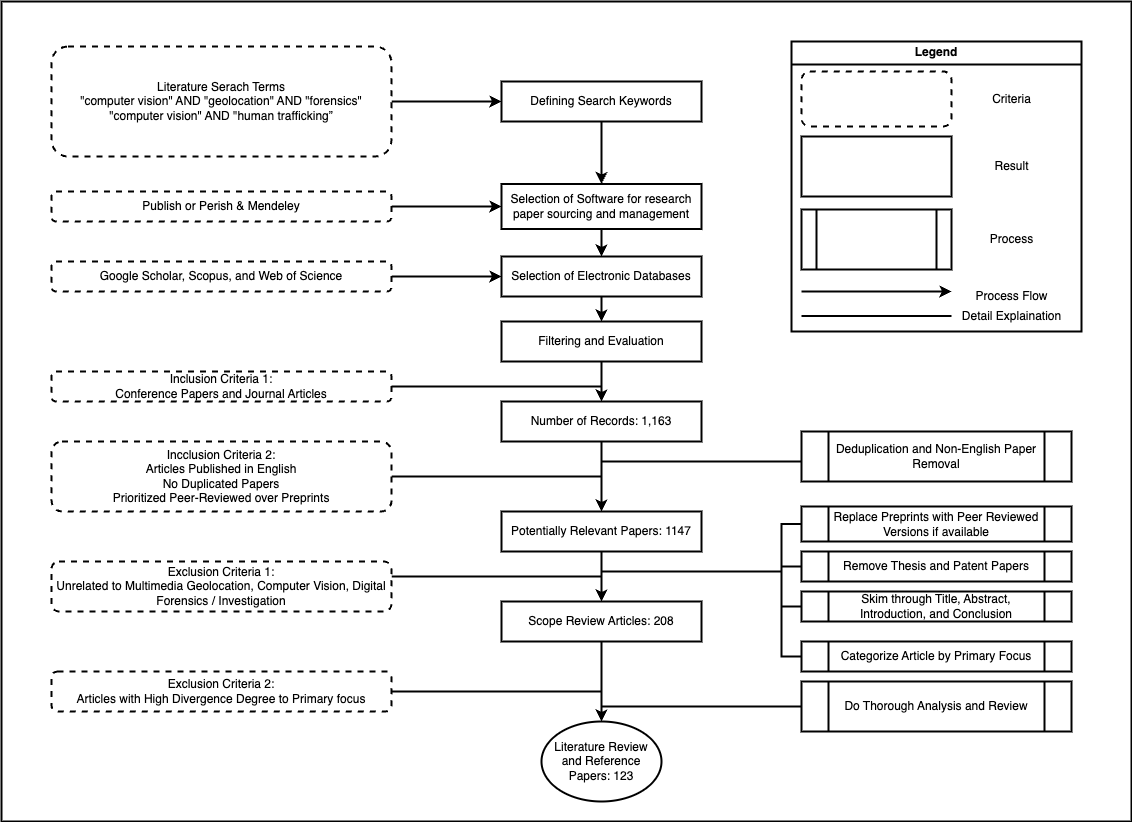}
  \caption{Procedures and Processes for Conducting the Systematic Literature Review}
  \label{fig:process}
\end{figure*}

\begin{itemize}
    \item \textbf{RQ1: What are the state-of-the-art computer vision techniques used in the context of multimedia geolocation in digital forensics?}
    \\As computer vision is a continuously expanding discipline with constant developments in techniques and algorithms, understanding the cutting-edge approaches used in the sphere of digital forensic multimedia geolocation is critical. This information will provide insights into the current cutting-edge methodologies and technologies being utilized to solve geolocation difficulties in digital forensics, as well as assist in identifying gaps or chances for additional research and development.
    \item \textbf{RQ2: How is computer vision contributing to efforts aimed at countering human trafficking?}
    \\This question focuses on how computer vision techniques contribute to the efforts aimed at human trafficking combatting. Understanding the capabilities of these techniques may improve fundamental features of multimedia geolocation, allowing us to obtain insights into the strengths and limits of existing systems, and guide the development of new approaches for more accurate and efficient multimedia evidence analysis in human trafficking investigations.
    \item \textbf{RQ3: What are the potential implications of multimedia geolocation for human trafficking investigation and prosecution?}
    \\This research question underscores the pivotal role of geolocation in modern human trafficking investigations. By examining the multifaceted implications, we can determine how geolocation enhances the collection, authentication, and presentation of digital evidence that could be a breakthrough in victim identification, tracking trafficking routes, and strengthening cases against traffickers.
\end{itemize}

\subsection{Data Source Selection and Conducting Searches}
To ensure a concentrated and complete examination of literature relating to the intersection of computer vision, forensics, and geolocation, the search phrases ``computer vision'' AND ``forensics'' AND ``geolocation'' were chosen. These terms were chosen for their relevance to the research topic and the goal of locating scholarly articles that expressly address the use of computer vision techniques in forensic applications, with a focus on geolocation. To restrict the search and discover literature that mainly explores the use of computer vision in solving human trafficking issues, the search phrases ``computer vision'' AND ``human trafficking'' were chosen.

\indent\textit{Publish or Perish;} a software program for searching retrieves and analyzing academic articles and citations was used to conduct independent searches on a variety of electronic data sources such as Google Scholar, Scopus, and Web of Science. The searches were conducted on the 27th of January, 2023. The combined search results totaled 1,163 articles from which non-English, patent papers were removed, and non-peer-reviewed articles were replaced with peer-reviewed versions if available, or otherwise removed. We proceeded with article deduplication and analyzed the remaining articles. Figure~\ref{fig:process} depicts the processes involved in the systematic literature review.

\subsection{Sorting and Selecting Studies}
The next phase of the review was to properly sort the resulting articles for selected studies through a collaborative effort from the authors. This was achieved by using Mendeley; a bibliography management tool. The results of the initial searches are carefully analyzed in a multistep process. First, the article titles and abstracts were assessed to determine their relevance and compatibility with the research topic. This phase aids in quickly eliminating irrelevant articles and limiting the selection. The remaining articles' conclusions or summary sections were then studied to acquire a better understanding of the results and contributions of each study. Finally, the full-text papers are thoroughly reviewed and studied to examine the methodologies, results, and implications. All eligible and valuable papers for inclusion in the research on computer vision for multimedia geolocation in human trafficking investigation were identified by following this systematic approach of title and abstract reading, conclusion reading, and full-text article reading.


\subsection{Data Extraction and Synthesis}
The sorted and selected articles were synthesized through relevant data extraction methods to form a well-detailed review. This process started with article categorization in Mendeley where each article was annotated between the review topics that form part of this paper. Articles were categorized based on: 1) The adopted computer vision techniques [image segmentation, object detection, classification, or mixed], 2) Multimedia data type [text, image/video, sound,  or multimodal], 3) Relevance to digital forensics, multimedia geolocation, human trafficking, hotel room classification, and law enforcement on a scale of high, medium, low and none, indicating irrelevancy to areas of interest. After a detailed analysis of each article, several were deemed irrelevant based on the aforementioned criteria and categories. This resulted in 123 articles remaining for inclusion in the final systematic review. The authors proceeded to write a summary of these relevant papers to have a synthesized review that provides answers to the above-stated review questions. Table A.1 in the appendix of this paper provides a full overview of each article included as part of this evaluation, and their respective classifications.

\section{Background: Human Trafficking}
\label{sec:HT}
Human trafficking (HT) is a complex and pervasive global problem that involves the exploitation of people for a variety of reasons, including forced labor, sexual exploitation, and involuntary servitude for monetary gain or other benefits. It is a type of human slavery that affects people of all sexes, ages, cultural backgrounds, religions, socio-economic classes, education, and geographical locations~\cite{Goh_Findind_2021, Karnik_A-new_2022, dasgupta2022audio, Vinavatani_AI_2022}. Since the Palermo Protocol to Prevent, Suppress, and Punish Trafficking in Persons was ratified in 2000, there has been a growing awareness of HT within scholarly communities and the general public~\cite{sierra2022twitter, Deeb-Swihart_Ethical_2022}. Although precise data are difficult to come by, in 2016, the United Nations International Labour Organization estimated that 40.3 million people were in forced labor~\cite{dasgupta2022audio, Li_Detection_2018}, and in 2017, the Global Estimates of Modern Slavery Report estimated that HT over 25 million people and earns more than \$150 billion US dollars in illicit profits annually worldwide~\cite{dimas2022_Operations-research, Kshetri_regulatory_2023}. Additionally, the Polaris Project predicted that these operations will increase by 20\% in 2018–2019~\cite{Upadhayay2021CombatingHT}.

The prevalence of escort websites, where human traffickers can openly advertise amid at-will escort advertisements, has contributed to the significant increase in sex trafficking, the predominant form of human trafficking~\cite{tong_combating_2017, Paul_Multi-modal_2020}. More than 4 million children worldwide fall victim to child trafficking each year, often for the purpose of child sexual exploitation~\cite{Mukherjee_A-Machine_2021}. The global sex trade is home to over 2 million children who are used as prostitutes, with ages ranging from 11 to 14 on average, according to the United Nations Children's Fund (UNICEF). Between 2009 and 2012, there was a 25\% increase in the proportion of children (between the ages of 2 and 18) who were victims of trafficking~\cite{Hemadri_ChildrEN-SafEty_2022}. Also, approximately 800,000 children are reported missing each year in the United States, according to the National Centre for Missing and Exploited Children website (NCMEC), and 250,000 children are reported missing each year in the European Union (EU) according to Missing Children Europe (MCE) and the European Federation for Missing and Sexually Exploited Children~\cite{Koudelová_Modelling_2014}.

In digital forensic labs all over the world, child exploitation investigations are one of the more popular investigations kinds. The growing use of anonymization software, private P2P networks, and cloud-based Kernel-based Virtual Machine systems has made these analyses difficult. This move has enabled traffickers to post their victims in numerous geographic locations at the same time, while also boosting operational security through the use of multiple channels of electronic communication with purchasers; complicating law enforcement's efforts to break these unlawful organizations~\cite{Li_Detection_2018}. The fight against human trafficking and the use of child sex exploitation materials has been carried out by law enforcement agencies worldwide from different perspectives~\cite{Anda_Improving-Borderline_2019, Swaminathan_Predict_2022}.

The crime of trafficking in persons presents itself as a sequence of subsequent actions rather than a single violation with regard to the manner of operation. The four steps of the process include victim recruiting, victim transportation, victim exploitation, and management of later criminal gains~\cite{Raets_Trafficking_2021}. Traffickers market the criminal services of the people they supply through Internet classified advertisements, bulletin boards, and social media \cite{Sethi_Large-Scale_2013} but web-based HT activities are still sporadic among escort adverts and challenging to spot because of its often-latent nature~\cite{Hundman_Always-Lurking2018}. One of the driving forces behind this was the extensive use of online sex reviews and adverts, even on the Open Web. As a result, law enforcement operations to prevent minor human trafficking strive to leverage the vast information source that advertisements provide as a source of evidence~\cite{Stylianou_TraffickCam_2017}. Building intelligent systems for supporting investigators and facilitating counter-human trafficking activities has arisen as an essential objective in light of the alarming increase in online and offline sex advertisement activity, of which a sizeable fraction may be related to trafficking.

\begin{figure*}[h!]
\centering
\includegraphics[width=17cm, height=7cm]{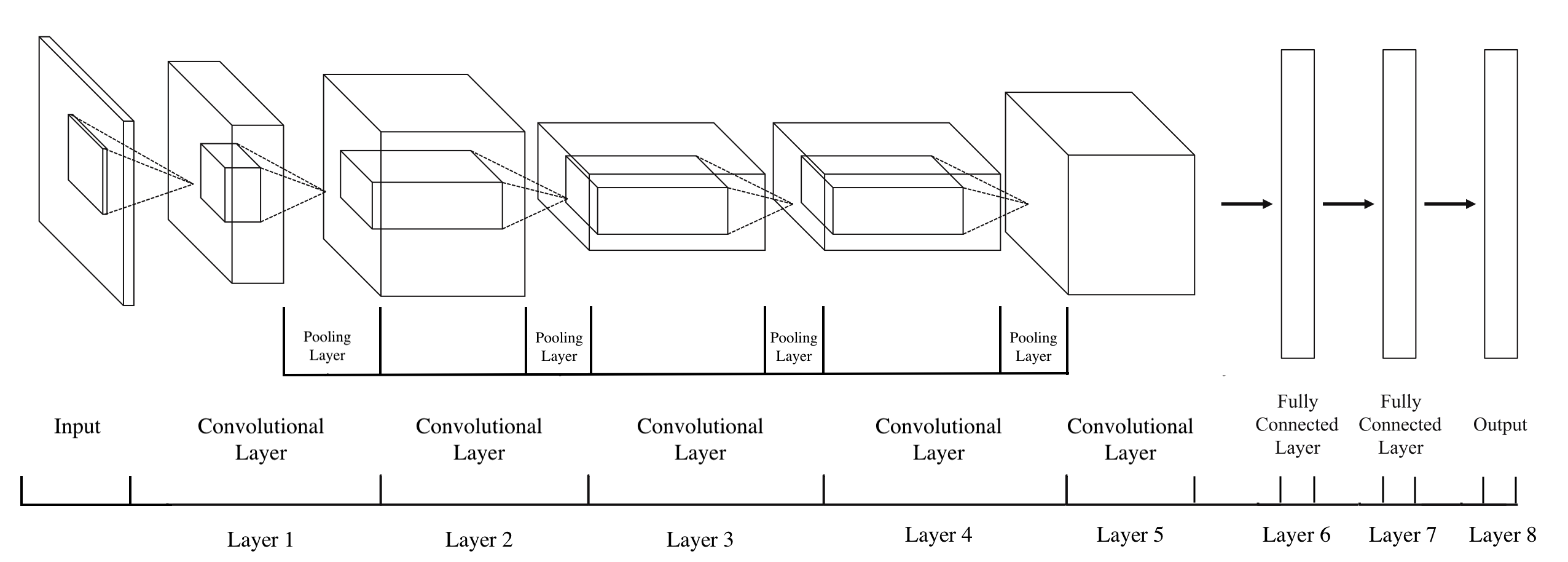}
  \caption{Convolutional Neural Networks (CNN) architecture~\cite{Piras_Information_2017}}
  \label{fig:CNN}
\end{figure*}

\subsection{Combating Human Trafficking}
Due to the enormous scope of HT activity and the variety of ways that people might be taken advantage of, it takes great effort from different disciplines and viewpoints to comprehend and confront such a complicated crime~\cite{Goh_Findind_2021, Koudelová_Modelling_2014}. The lack of actionable intelligence concerning trafficking incidents and operations presents a significant obstacle in the fight against worldwide human trafficking~\cite{Upadhayay2021CombatingHT}. From an analytical perspective, methods such as general Machine Learning, (classification and clustering), Semi-Supervised Learning, Natural Language Processing, Computer Vision, and Search and Retrieval are proven useful among researchers while Operation research methods are the default for law enforcement agencies~\cite{dimas2022_Operations-research}. Artificial intelligence has transformed criminal investigation and tracking. This can be used by investigators to teach their systems how to recognize parts of crimes from sources such as social media or surveillance film~\cite{Horan_cyber_2021}. Thus, the application of intelligent systems to identify human trafficking can directly affect the allocation of investigative resources and decision-making, as well as, more widely, help address a pervasive social issue.

\citet{Goh_Findind_2021} explored if crowdsourcing may be used to fight child trafficking because kids are frequently moved across borders or to various sections of a country. \citet{Hundman_Always-Lurking2018} in their work, approached this problem as a classification task by developing an HT detection system capable of classifying online adverts as trafficking related or not. Likewise, \citet{Upadhayay2021CombatingHT} developed an automated pipeline based on recent developments in natural language processing and machine learning to streamline the curation, analysis, and extraction of useful intelligence from a variety of news sources to address this issue. \citet{Mukherjee_A-Machine_2021} created a machine learning proof-of-concept pipeline to help identify minors in intercepted photos. Two machine learning models make up their suggested pipeline: 1) determining whether a picture of a child has a school uniform, and 2) identifying characteristics of various school uniform pieces (such as the color/texture of shirts, sweaters, blazers, etc.). \citet{tong_combating_2017} created a multimodal deep model termed the Human Trafficking Deep Network to identify human trafficking advertisements in the Trafficking-10K dataset. \citet{Hemadri_ChildrEN-SafEty_2022} instead, approached this problem as a search and retrieval task by developing a system that uses Memory Augmented ScatterNet ResNet Hybrid network to encode the input image of a victim and then searches the encoding with the Age, Kinship, and Sketch databases to determine the victim's identity. \citet{Karpagam_A-novel-2022} proposed a facial recognition system that is focused on deep learning to find people who have gone missing due to HT and saving them by identifying them with the help of modified deep belief networks and Horn Schunck optical flow motion estimation.

Automatic trafficking identification is a crucial application of AI for the benefit of society. Still, it also teaches us to be cautious when using predictive machine learning algorithms without reliable debiasing~\cite{Zavrsnik_criminal_2020}.

\section{Background: Computer Vision}
\label{sec:backgroundCV}
Computer vision is a branch of artificial intelligence concerned with enabling computers to interpret and comprehend visual information from the outside world, such as images and videos, through learning algorithms. It employs a variety of techniques and methodologies to analyze, process, and extract useful insights from visual data, ranging from traditional computer science methods to more contemporary deep learning techniques. This can involve but is not limited to detecting specific objects, segmenting (separating) various objects within a scene, tracking these objects over time and space, three-dimensional reconstruction of the objects and/or scenes, and determining the camera's position in the scene over time and space and many more~\cite{aronson_computer_2018}.

\subsection{Deep Learning and Neural Networks} 
This constitutes a branch of machine learning in artificial intelligence, and a modern approach to computer vision, using multiple-layer neural networks to extract high-level features from digital data~\cite{iqbal_machine_2020}. Deep learning has produced and substantially enhanced results in a variety of computer vision applications, including image classification, single and multiple object detection, picture captioning, and face recognition~\cite{Ivasic-Kos_Application_2022, Yokota_Visual-based_2020, Matasci_Deep_2021}. With the approach, latent features and semantic cues can be extracted for retrieval, classification, and other modes of geolocating multimedia contents \cite{Nam_Revisiting_2017, Jiang_Hierarchy-Dependent_2019}.

Neural networks are a fundamental component of deep learning, as deep learning algorithms use interconnected layers of neurons to process, transform, and learn from complex data. These computational models are inspired by the structure and function of the human brain to process information~\cite{Keerthi_RIPA-2018}. The neural networks can be cascaded and arranged in different ways to form a suitable architecture for a proposed problem. These architectures, such as convolutional neural networks (CNNs) are made up of several layers of convolutional and pooling algorithms, as shown in Figure~\ref{fig:CNN}, that capture various degrees of abstraction. Recurrent neural networks (RNNs) and CNNs algorithms have transformed computer vision tasks such as feature extraction from images and videos~\cite{Nilwong_DeepLearning_2019}.

In the low-level feature extraction process, convolutional operations traverse input matrices with convolutional kernels that serve as filters to automatically learn parameters like kernel size, strides, channels, padding, and activation via deep learning method. The result is then subsampled using a layer pooling technique to discover high-level features that represent the original input~\cite{Ammatmanee_Transfer-learning_2021}.

Deep learning algorithms are also used in localization and positioning applications. For example, the deep learning-based encoder predicts picture locations based on low-level characteristics as seen in \citet{Nilwong_DeepLearning_2019}. Also, \citet{Zhu_Large-Scale_2022} used a collection of neural networks and algorithms to extract important attributes about panoramic imagery for architectural analysis and localization. \citet{Jiang_Hierarchy-Dependent_2019} extracted semantic visual cues, especially objects and key frames from videos, for venue category prediction. \citet{Keerthi_RIPA-2018} used two multi-layered CNNs; the first as transfer learning and the other specifically trained from scratch to classify over 81 different classes of images for real-time image privacy alert systems. The authors noted that using supervised learning, a deep convolution neural network can produce excellent results. In addition, \citet{Neal_Global_2019} used deep learning to build a more accurate geolocation model from microbiome data for identifying the provenance of material at a crime scene. The authors demonstrated the utilization of DeepSpace to estimate the surface intensity of a spatial point pattern using an ensemble of deep neural network classifiers.

\subsection{Computer Vision Techniques Related to Multimedia Processing}
Despite the fact that the content of an image may be represented by traditional features like color, texture, shape, scale-invariant feature transform (SIFT), or spatial positions, artificial intelligence-based features are increasingly being considered by academia and practitioners for image classification tasks among others. This is because the artificial neural network (ANN) algorithm can mimic human cognitive processes and acquire training to classify an image through deep learning with less labor expense and human error~\cite{Ammatmanee_Transfer-learning_2021}.

Computer vision techniques such as object detection, feature extraction, and image segmentation among others can be used to identify specific objects, scenes, or patterns in the multimedia data \cite{Salem_Learning_2020, aronson_computer_2018}. When integrated with machine learning, computer vision techniques can be used to detect objects in digital photographs and rebuild scenes from images~\cite{aronson_computer_2018}. This information can then be utilized to estimate the location of the multimedia or to detect correspondences, which could range from object matching to comparing the style between images and the semantics of the two~\cite{moreira_image_2018}. A number of these techniques are outlined in the following sections.

\subsubsection{Feature Extraction} 
The process of identifying and extracting relevant features from an image or video stream for subsequent analysis such as search, retrieval, and classification~\cite{Zhu_Large-Scale_2022, Black_Evaluation_2020}. Specific descriptions that objectively extract visual elements from image content are frequently employed to quantify the similarity between pairs of images. We can extract two distinct kinds of features from the visual contents, depending on the application. They are described by both local and global elements. Most times, local features are used for higher-level applications like object recognition, whereas global features are used for low-level applications like item detection and categorization. The mix of global and local characteristics can improve accuracy while raising processing costs.

These descriptions are known as visual descriptors in the field of computer vision, and they are used by various SIFT and speeded-up robust features (SURF) for color content comparison, texture components comparison, and Haar feature detection among many others~\cite{Yokota_Visual-based_2020, Smelyakov_Search-by-Image_2018}. Local and global descriptors have the power to draw attention to image blocks that exhibit extremely distinctive responses from features in the image. The aforementioned characteristics will, however, highlight the most dominating target and stifle the others if prospective targets are provided in more than one block. \citet{Makantasis_Semi-supervised_2015} made use of a window descriptor to compare each picture block with its surrounding blocks in order to solve this issue. This was also of interest to \citet{Yadav_A-Feature-Averaging_2019}. For each facial image in their work, the authors used several feature descriptors to produce a high-dimensional feature vector before using the Principal Component Analysis to reduce the dimension.

Deep learning techniques are widely employed for feature extraction in computer vision tasks. This is because raw data are processed by the network, which provides an internal feature representation of the data suitable for the task at hand. These networks stand out at this task because of their capabilities to extract these features from scratch without any prior hints or knowledge~\cite{Piras_Information_2017}. In addition, CNNs are a common technique for automatically learning hierarchical representations of visual data from images~\cite{Wu_Leveraging_2022, Liu_Age-Invariant_2022}. This was seen in \citet{Jiang_Hierarchy-Dependent_2019} work where the authors proposed a Hierarchy-dependent Cross-platform Multi-view Feature Learning framework based on a CNN architecture for predicting venue categorization from videos using pictures from other platforms.

Similarly, \citet{Yokota_Visual-based_2020} adopted DEep Local Feature (DELF) which is a deep learning-based descriptor to detect and extract local features attributed to the same object instance in their development of a visual-based geolocalization framework. The authors used two steps to test if DELF descriptors can extract specific aspects from image content for classification. The first step used the t-distributed stochastic neighbor embedding approach, to compute the dissimilarity matrix, which contains the dissimilarities between all pairs of photos in the Tokyo image collection, and show it in two-dimensional space. The second step determined the mean average precision (mAP) and recall, which are commonly employed in information retrieval systems to quantify how well a system can anticipate or retrieve user-required items from a database. In the context of geolocalization, mAP indicates how closely the area from which the searched top-n photos were obtained corresponds to the area from which the query image was taken.

RNNs are another technique effective for sequence data analysis, such as processing sequential images or videos. RNNs are capable of capturing temporal dependencies and extracting characteristics over time. Furthermore, CNN variations such as Residual Networks (ResNets) and DenseNet incorporate skip connections or dense connections between layers to increase feature learning and model performance. These architectural deep learning algorithms enable successful feature extraction from multimedia.

\subsubsection{Object Detection} 
This is the process of identifying and localizing objects within an image or video stream. This technique can also be used for three-dimensional reconstruction of the object or scenes~\cite{aronson_computer_2018}. It is a fundamental computer vision technique that encompasses object localization and classification and is referred to as a regression job. In addition to detecting objects within images, it is also sometimes used to recognize the same object during a video or in distinct footage, known as object reidentiﬁcation. This can be very challenging and error-prone due to context-related issues like noise, occlusion, varying illumination, ambiguity, and moving background objects~\cite{Kraus_Toward_2020}. Object detection has recently made considerable advancements, riding the tide of convolutional neural network (CNN) architectures because of its ability to handle visual input~\cite{Nilwong_DeepLearning_2019}. Also, since AlexNet won the 2012 ImageNet competition, object identification algorithms based on CNN quickly advanced to the top of the object recognition field with good performance despite the lack of strong mathematical arguments~\cite{Xiao_Tiny-object_2023}. As such, there are groundbreaking frameworks for achieving this goal in different ways such as Mask R-CNN, Fast \& Faster R-CNN, and other variants that use a region proposal network (RPN) to generate object proposals (potential bounding boxes) and subsequent classification network to classify them~\cite{Zhu_Large-Scale_2022}. The Region Proposal Network (RPN), which decreases Faster R-CNN detection time and boosts accuracy, is a significant distinction between Faster R-CNN and Fast R-CNN. Faster R-CNN for Object Detection is appropriate for real-time applications because of its enhanced speed. Another real-time object detection framework that uses a single neural network architecture to generate object proposals and classify them in a single pass is the Single Shot MultiBox Detector (SSD).

Several recent publications utilize YOLO (You Only Look Once), which is a real-time object detection framework renowned for its speed and accuracy. Its architecture is based on a single neural network that predicts bounding boxes and class probabilities straight from complete images \cite{Yuan_geolocation_2022}.
The success of this technique was seen in many works, including \citet{Kraus_Toward_2020} where the authors used a pre-trained YOLO v3 module in combination with a state-of-the-art re-identification approach in their video analysis. Also, with a deep learning-based object detection method, \citet{Matasci_Deep_2021} used RetinaNet to identify vessels in a big collection of optical remote sensing images.

\subsubsection{Image Segmentation} 
The division of an image into many segments or regions having comparable properties~\cite{aronson_computer_2018}. Image segmentation is also a critical stage in many computer vision applications, such as object recognition and scene comprehension~\cite{Karnik_A-New-Deep-Model_2022}. By dividing an image into sections with comparable qualities, such as color, texture, or intensity, image segmentation aims to simplify the representation of an image. This approach paves the way for more sophisticated analysis and interpretation by enabling computers to recognize and comprehend various items and structures within an image.

This has received attention from researchers, including \citet{Yuan_geolocation_2022} who used this to geolocate visual contents taken indoors. Through this technique, the author was able to measure the spatial patterns in indoor images. Similarly, segmentation was also one of the crucial factors in the success of \citet{workman_revisiting_2022} on remote sensing with geospatial attention, where the authors considered using overhead imagery segmentation to localize ground-level images. \citet{Karnik_A-new_2022} in their work on image classification employed the ``self-correction for human parsing'' model to segment regions of interest in facial images.

Segmenting images in multimedia makes it simpler to spot and identify items, people, or landmarks, thus improving the precision and efficacy of forensic and geospatial investigations.

\subsubsection{Feature Matching and Tracking} This involves locating and contrasting shared elements or focal areas in two or more photos. These characteristics may consist of points, corners, edges, or other distinguishing visual components. The goal of tracking is to foresee the object's states in consecutive frames given the initial state of the item at the first frame, such as position and extent. This issue is challenging because of a number of elements, including occlusions, scale fluctuation, deformation, backdrop clutter, illumination variance, and quick motions.

By establishing correspondences between characteristics in various images, activities like image alignment, image stitching, object detection, and 3D reconstruction of images can be made possible. These can be feature-based matching, optical flow-based tracking, and point cloud matching, which are used in photos or videos to track objects or features of interest over frames or scenes~\cite{Shao_Can-We-Track_2019}.

Some research works that rely on this technique include \citet{Shao_HRSiam_2021} who developed a compact parallel network with a high spatial resolution to match and track tiny objects in satellite videos. The authors noted that the modern cutting-edge tracking techniques virtually use the inter-frame motion information that films naturally include, instead focusing mostly on high-level deep features of a single frame with limited spatial resolution. With their architecture, real-time and precise localization when applied to the Siamese Trackers is guaranteed.

\subsubsection{Image Enhancement}
Denoising, deblurring, and color correction are examples of techniques used to improve the quality and clarity of visual data, particularly under difficult settings such as low light, motion blur, or noise \cite{aronson_computer_2018}. This is a very crucial task in computer vision task because visual content often requires some sort of enhancement for the algorithm to properly capture features of interest. This was seen in \citet{Aspandi_Heatmap-Guided_2019} where the authors reduced their dataset variability through a deep learning-based image denoising autoencoder.

Human observers or automated systems can comprehend and evaluate the multimedia with less ambiguity when hidden details are revealed, contrast is improved, noise is reduced, and certain characteristics of the image are highlighted.

\subsubsection{Transfer Learning}
This is a technique widely used in deep learning and  computer vision. It uses pre-trained models from large-scale datasets to solve similar tasks with little labeled data. In place of building a CNN or other artificial intelligence feature extracting algorithms from scratch, the transfer learning technique applies knowledge learned from one task to improve the performance of another~\cite{Ammatmanee_Transfer-learning_2021}. Transfer learning allows for the effective training of models on smaller datasets by transferring the learned features or parameters/weights from a pre-trained model that extracts low-level features such as patterns, edges, corners, and gradients (a general feature to visual contents) thereby reducing the requirement for substantial labeled data~\cite{Anda_Improving-Borderline_2019}. This method not only saves computational resources but also enhances model generalization and accuracy in a variety of computer vision applications, including multimedia geolocation.

Depending on the problem space, the effectiveness of a pre-trained model is sometimes dependent on the correlation between the trained dataset and the problem space dataset. While there are several pre-trained models built for different purposes, it is worth mentioning some that use deep convolutional
neural network models such as AlexNet, Inception V3, GoogleNet, SqueezeNet and ResNet50, VGG16, and many more~\cite{deshmukh2022human,DBLP:conf/iciot2/YangHPXL20}

\citet{Jiang_Hierarchy-Dependent_2019} leverage this technique to develop cross-platform transfer deep learning, capable of reinforcing the learned deep network from videos with images from other platforms. To begin, the authors trained a deep network (ImageNet to extract the visual object features) using videos with the aim of filtering images from different platforms, and these selected images are then fed back into the learned network (Places CNN to extract visual scene features) to improve it. In parallel to this, \citet{Keerthi_RIPA-2018} took advantage of this technique to improve the classifier feature learning in their real-time image privacy alert system development. Similarly, \citet{Ammatmanee_Transfer-learning_2021} utilized this technique to test and evaluate the potential of 11 pre-trained CNN for hostel image classification, from which the authors concluded that transfer learning can provide good performance on classification tasks with less computational cost.

\subsubsection{Embedding} Embedding is a powerful computer vision approach that allows high-dimensional data, such as images or videos, to be transformed into compact and meaningful representations in a lower-dimensional space, easing tasks such as similarity matching, clustering, and classification. The method employs an unsupervised forward pass through a neural network to construct a fixed-length representation. These embeddings capture semantic information as well as spatial linkages, enabling rapid and effective geolocation analysis of multimedia content.

\citet{Stylianou_TraffickCam_2017} created embeddings to differentiate between images in the collection of crowdsourced hotel room images in the fight against human trafficking. Also with this technique, \citet{Zhu_Large-Scale_2022} generated a 4096-dimensional embedding to extract deep features from Panoramic imagery for the purpose of distribution visualization and performing similarity search. \citet{Matasci_Deep_2021} in their work, to determine if two images represent the same vessel used a Twin neural network to encode these images into 
a low-dimensional embedding space. The authors noted that the proximity of the two images to one another can be calculated using these embedding vectors to produce a distance or similarity metric.

With the use of embeddings, the geolocation system may find commonalities across various multimedia files, assisting in the identification of objects, landmarks, and geographic places and facilitating more accurate matching and geolocation of multimedia data. In addition, embeddings enable cross-modal analysis, enabling the fusion of data from diverse sources, such as images, text, and metadata, to improve the precision and robustness of the geolocation process.

\subsection{Multimedia Analysis with Computer Vision}
Multimedia analysis with computer vision has evolved as a valuable tool in a variety of sectors, including the investigation of human trafficking. Hidden information in data can be spotted using visualizations without a specific formulation of a hypothesis~\cite{Kraus_Toward_2020}. Computer vision approaches enable the identification and categorization of objects, sceneries, or patterns within multimedia content from the standpoint of recognition~\cite{Black_Evaluation_2020}. Recognition algorithms can automatically detect and distinguish specific items or events of interest in assessing visual features and, utilizing machine learning techniques, assisting in the detection and understanding of suspected trafficking-related aspects.
 
Convolutional neural networks (CNNs) are one of the many types of ANNs that have been used most frequently in well-known image categorization projects including AlexNet, ZF Net, GoogLeNet, and ResNet. With classification techniques, the categorization and organizing of multimedia data based on predetermined classes or criteria allow for more effective data management and retrieval. For instance, \citet{Aspandi_Heatmap-Guided_2019} in their work trained the state-of-the-art Squeeze Excitation Network for family classification in the wild. \citet{Ammatmanee_Transfer-learning_2021} leverages transfer learning for the classification of hostel images, thereby reducing if not eradicating human errors in achieving the same goal. 
 
Furthermore, search and retrieval technologies enable efficient exploring and retrieving of relevant multimedia content based on specific search queries or similarity metrics. While there are several ways to effectively develop a search system such as text-based and content-based, the content-based analysis with computer vision seems to be more accurate on the retrieves.
 
Because images are searched on localized pages, text-based search is dependent on the text in which images are described, hence the retrieval result is dependent on the language used to post the query. These early discoveries paved the way for researchers to investigate content-based image retrieval (CBIR) systems based on the fusion of different features, either by employing weighted similarity measures, in which different features are weighted based on their relevance to the task at hand or by combining different similarity functions, in which image similarity is computed separately for each feature and then their values are combined. The origins of this technique can be found in the fact that the performance of a CBIR system is fundamentally limited by low-level features, and therefore cannot provide adequate retrieval results when users' high-level notions are not easily conveyed by low-level features~\cite{Piras_Information_2017}.

In the early 1990s, the computer vision community began to investigate CBIR. Intuitively, the basic aim of the CBIR system is to reduce a huge set of photos to a small subset of those with the needed attributes by exploiting the visual content of the image. \citet{kato1992database} coined the term CBIR in 1992 to describe an experiment to automatically search photographs from a database using color and form representations. The most frequent method for searching for images by image content is for the user to input a query image, and then the system to find images that are identical or similar to the query images.

Considering that an image is made up of non-meaningful image elements (pixels), establishing the visual image attributes utilized in the search is the first stage in locating or searching for images by content. Low-level elements such as color, shape, texture, or any other information derived from the image itself are used to represent an image's visual content. Previously, several techniques such as color histograms, edge, interesting point detectors, gradients, and so on were utilized, but now the features are created automatically utilizing deep neural networks and computer vision. The foundational study, in which CNN was utilized in an image classification challenge, clearly shows that characteristics arising in the upper layers of the CNN can also serve as good image retrieval descriptors~\cite{Piras_Information_2017}.

While these search and retrieval results can often yield positive results, it is important to have a complementary provenance analysis unit that can validate the search result against desired realistic scenes and distinguish the image of interest from other similar images. This could simply be through metadata verification of searches as seen in \citet{moreira_image_2018} work.

\begin{figure}[H]
\centering
\includegraphics[width=0.48\textwidth]{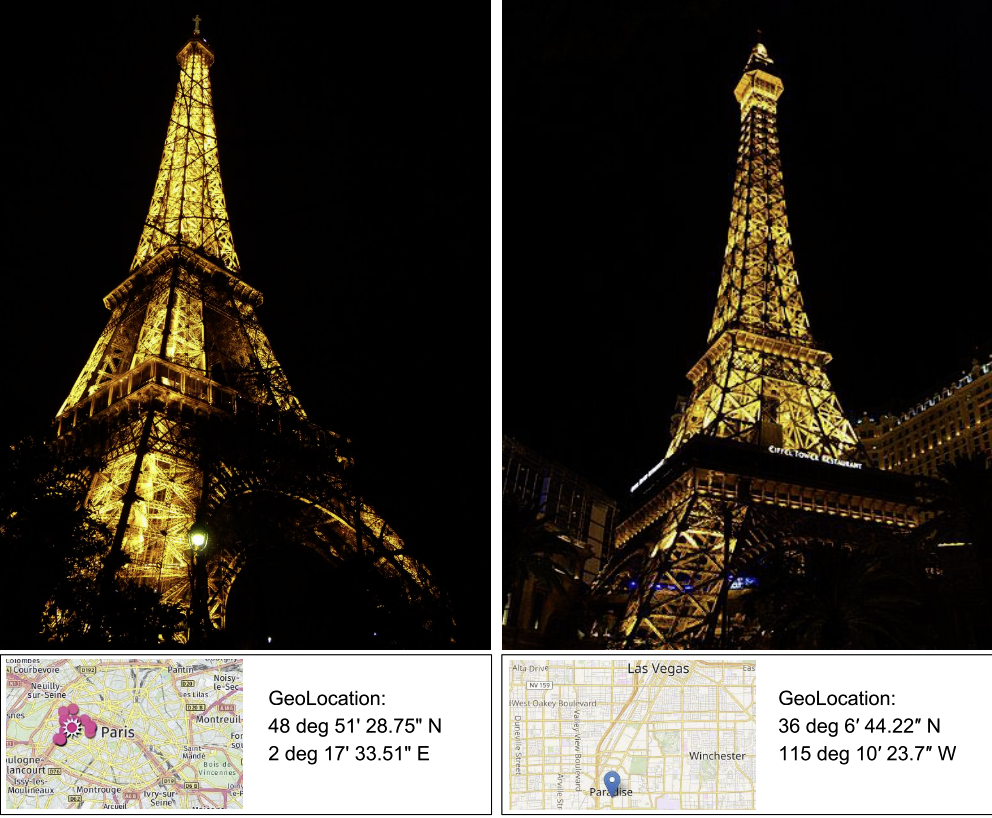}
  \caption{Similar images based on visual content but with differences in metadata~\cite{moreira_image_2018}}
  \label{fig:imageProvenance}
\end{figure}

As seen in Figure~\ref{fig:imageProvenance}, Left; A nighttime photograph of the Eiffel Tower in Paris.
Right; A nighttime photograph of the Las Vegas reproduction of the monument. Both photographs portray identical visual items — only the image file metadata can tell us that they are from different locations.
 
\subsection{Computer Vision for Multimedia Geolocation Estimation}
Multimedia geolocation is the process of estimating the real-world location in which a multimedia file, such as an image or video, is taken~\cite{Glistrup_Urban-Image_2022}. Similarly, image geolocalization requires the ability to extract visual attributes from images and an understanding of the geospatial distribution of these attributes \cite{Salem_Learning_2020}. This task is becoming increasingly important in digital forensics as the number of multimedia files used in criminal investigations such as human trafficking increases, and many of these files lack proper geographical annotations or tags~\cite{Sethi_Large-Scale_2013, Ivanov_Comparative-Study_2013, Nam_Revisiting_2017}.
The complexity of geolocating, geotagging, or finding geographical clues to these contents increases with a decrease in contextual references when analyzing indoor scenes as against outdoor scenes~\cite{Yuan_geolocation_2022, Friedland_Multimodal_2010}.

Computer vision techniques offer exciting potential for accurately evaluating and interpreting visual content, such as photographs and videos thereby, improving the precision, accuracy, and reliability of multimedia geolocation estimation~\cite{Glistrup_Urban-Image_2022,moreira_image_2018} with effective approaches to detecting human trafficking events~\cite{dasgupta2022audio}. Due to the recent advances in computer vision and deep learning, we can now acquire more specific insights about the context of an image and the interactions between items inside it, allowing us to better handle these difficulties~\cite{Keerthi_RIPA-2018}. The analysis of multimedia data within the context of geolocation entails the use of computer vision techniques to process and interpret these various types of media, allowing for the accurate identification and localization of relevant objects, landmarks, or events to aid in investigative efforts but at the expenses of computational resources and cost~\cite{Murdock_2016}.

While various techniques sometimes have vastly varied scopes: although some systems cover the entire globe, the majority operate on a much smaller scale, such as inside a single city. It is important to note that, due to a lack of standards for assessing geolocalization algorithms and systems, it is currently difficult to directly compare the various techniques. They not only cover extremely various geographical scopes, but they also operate on very distinct datasets and are assessed using a number of different task-specific criteria.

As there is no direct approach to how multimedia content can be geolocalized, several researchers have attempted this problem using different techniques to give the closest location estimation. Early approaches typically treat this as a similar picture-matching problem, employing geotags extracted from matched database photos. Recent research has looked into how to estimate finer-grained and more extensive geo-context information, such as viewing direction estimation~\cite{Choi_Multimodal_2015}. These techniques can be categorized into three major groups; metadata analysis such as image title, metadata; tags, EXIF (Exchangeable Image File Format) which is the most popular format according to \citet{Friedland_Multimodal_2010,lallie_geotract_2013}, and description, image or video content analysis, and the fusion of both data domains as multimodal data~\cite{Glistrup_Urban-Image_2022, Friedland_Multimodal_2010, Sethi_Large-Scale_2013}.

For multimedia content taken outdoors, several publications leverage the use of natural cues in the multimedia files. These cues include landmark detection, sun azimuth calculation, building and skyline detection, and many more. \citet{Ivanov_Comparative-Study_2013} worked on a system for automatic geotag propagation in images based on the similarity between image content, such as famous landmarks, and its context. The authors conducted a comparative study of five different trust modeling methods, including two previously proposed methods and three new methods proposed by the authors. \citet{li_validating_2018, Kakar_Authenticating_2012} in their experiment, relied on the law of  nature that sun position varies with the time and location to validate the integrity of image content which might have been altered. The authors propose a novel approach that utilizes image metadata, location, and semantic scene understanding to validate the contextual information of outdoor images by computing the sun's position using astronomical techniques such as the sun azimuth calculation. Details of these approaches are discussed in Section~\ref{sota_multimedia_geolocation}.

\section{State of the Art in Multimedia Geolocation} 
\label{sota_multimedia_geolocation}
The term ``multimedia'' in this context refers to the combination of different types of digital data formats, such as photographs, videos, audio recordings, and textual data~\cite{Worring_Multimedia_2016}. It includes a variety of media formats that are commonly analyzed in digital forensics and analysis, such as human trafficking investigations. Geospatial or geolocation data is any data that may be used to pinpoint a specific geographical place~\cite{lallie_geotract_2013}. Multimedia data has a wealth of information that can be used in geolocation activities to extract useful insights for determining the geographical origin, context, and connections associated with human trafficking. Geolocation can broadly be categorized into indoor and outdoor with respect to where they were taken~\cite{Yuan_geolocation_2022, Stylianou_Indexing-open_2015}.

\subsection{Indoor vs. Outdoor Scenes} Recent research has identified various localization strategies for both outdoor and interior context~\cite{Nilwong_DeepLearning_2019}. It is worthwhile to explore the distinction between indoor and outdoor scenes before deep-diving into the details. Indoor scenes are often contained in or within the bounds of a building, such as rooms, and corridors. On the other hand, outdoor scenes include open areas, roadways, and landscapes. Geolocating multimedia material in interior environments presents distinct obstacles, such as the limited availability of GPS data, the presence of occlusions, and fluctuating illumination conditions \cite{Choi_Invisible_2022, Yuan_geolocation_2022, Jiang_Hierarchy-Dependent_2019}. Outdoor scenes, on the other hand, benefit from the potential availability of accurate GPS metadata as well as a wider range of visual cues such as landmarks, street signs, and natural features \cite{Zhang_Landmark_2018, Kakar_Authenticating_2012, salem_timestamp_2022}. It is important to understand the characteristics, procedures, and limits involved with geolocating multimedia material in both indoor and outdoor settings. These are discussed in Section~\ref{sec:techniques}.

\subsection{Techniques for Geolocating Multimedia Content} 
\label{sec:techniques}
  There are several techniques involved in the geolocation of multimedia content based on different modalities, e.g., time of capture, availability of metadata and description tags, indoor scene, outdoor scene, etc. Some of these techniques use sequences of photos, the relation of ground-level views to aerial imagery, correlating video stream with satellite weather maps over the same period, leveraging of GPS-tagged images of urban environments, multiview geometry, and feature matching~\cite{Choi_Multimodal_2015}. Among these techniques, the most popular are classification and retrieval-based solutions, with different strengths and weaknesses.
  
  For classification, machine learning and deep learning algorithms are trained to categorize multimedia data into different geographical classes based on visual attributes. These algorithms learn patterns linked with diverse locations' landmarks, architectural styles, or environmental factors. With the classification approach, researchers often achieve higher accuracy, fine-grained geolocation, efficient processing, interpretability, and generalization~\cite{ Stylianou_TraffickCam_2017, Yuan_geolocation_2022, Friedland_Multimodal_2010, Claudia_Live-forensics_2016}.
  
  Retrieval-based solutions on the other hand use similarity-based search techniques and lazy learning algorithms to return geotagged multimedia data from a database that closely matches a query image. Relevant photos can be found by comparing the perceptual features or metadata of the query image with the reference data, providing valuable information about the likely location where the query image was captured. This approach performs well if there are images with fields in the database that considerably overlap the query image fields. The major advantages of this mode are flexibility in querying, scalability, adaptability, the discovery of unanticipated locations, visual similarity exploration, and flexibility in geolocation granularity~\cite{Sethi_Large-Scale_2013, Glistrup_Urban-Image_2022, Stylianou_Indexing-open_2015, Nam_Revisiting_2017, Piras_Information_2017}. 

While there is no general technique that can effectively geolocate these contents without underperforming in certain scenarios, it is worth demystifying some notable approaches underpinned by researchers to solve this problem, as discussed in the following subsections.

\subsubsection{Metadata Analysis} Metadata in multimedia files is the typical starting point for identifying where the files were taken. Among myriad other uses such as image retrieval, content describing and summarizing, and captioning to mention a few, but there are many problems attached to this information. This problem includes lack of proper or missing geographical annotation, altered information with or falsely claimed by forgers to achieve malicious purposes~\cite{Ivanov_Comparative-Study_2013, li_validating_2018, Kakar_Authenticating_2012, Padilha_Content-Aware_2022, Choi_Invisible_2022}. These problems can result in the misrepresentation of multimedia chronology or misleading location estimation where precise and accurate geolocation analysis is important. It is important to highlight research that focused on the validation and correction of the metadata and those that utilized it for geolocation estimation purposes. Metadata encoded into an image, such as a timestamp and a geotag, might contain sensitive information about the people or objects within it~\cite{Keerthi_RIPA-2018}. \citet{Murdock_2016} noted that advertisements featuring victims being trafficked by the perpetrator frequently contain textual cues, such as physical traits, e.g., hair, body type, race, ethnicity, etc. and location information e.g., city and state that can help law enforcement and non-governmental organizations identify victims and take action to help them. The authors developed a preliminary Image Space system that uses multimedia metadata, or data about the actual photos and video that are normally recorded as they are made. This made the information retrieving system more precise to query searches.

\citet{Ivanov_Comparative-Study_2013} emphasized the importance of metadata in digital images and addressed the problem of missing metadata by automatically assigning landmark tags, with an emphasis on trust modeling to ensure that landmarks are accurately and reliably tagged. With visual similarity and spatial proximity between images as the major techniques used, landmark images were automatically tagged accurately, which contributed to multimedia geolocation estimation. Also,~\citet{moreira_image_2018} highlights the importance of metadata in understanding the context and authenticity of images. The authors explore the use of metadata associated with digital images, including EXIF data, GPS information, camera details, timestamps, etc., through data mining and machine learning techniques to identify patterns and inconsistencies and establish the authenticity and source of the images, which is essential to the geolocation.

The problem with metadata was also of interest to \citet{Li_AreYouLying_2017, li_validating_2018} and \citet{Kakar_Authenticating_2012} where the authors focused on authenticating image metadata, particularly geolocation information and sun direction estimation in outdoor images. \citet{Padilha_Content-Aware_2022} recognize that temporal metadata, with a special focus on the timestamps, can be manipulated to misrepresent the chronology or origin of multimedia content, which poses a significant challenge in geolocation analysis and investigation. The authors developed a content-aware approach to detect and identify such manipulations in the temporal metadata of multimedia files by analyzing the content, extracting relevant features, and comparing them with the associated timestamps to detect anomalies.

Knowing the time at which multimedia was taken can contribute to knowing where it was taken. It can provide useful temporal context, allowing investigators to link the multimedia to specific events or situations, i.e., if a multimedia file is timestamped during a known event or gathering, it can help narrow down the possible location. This argument is supported by \citet{salem_timestamp_2022} in their experiment to estimate timestamps from outdoor scenes using a learning-based approach through deep learning by analyzing various visual cues such as lighting conditions, shadows, and environmental factors. While images are handled in real-time by metadata connected with the image, e.g., timestamp, geolocation, tags, etc., retrieving them from an archive or geolocating may be difficult. This is because an image has rich semantic content that extends beyond the description supplied by its metadata \cite{Piras_Information_2017}.

\subsubsection{Pixels Contextual Analysis (Image or Video Content)} Image geolocation based on perpetual content has garnered much interest \cite{Piras_Information_2017}. Although metadata forensics alone were sufficient, but actual pixel (gradient) analysis is now required in order to produce more insightful findings\cite{Mattmann_Scalable_2017}. It is possible to match photos with reference databases or known geolocated images by detecting and extracting items, landmarks, scenes, and other contextual information within them. This information is crucial for geolocation, allowing investigators and researchers to determine the image's geographic context.Color distribution, texture, and shape properties in a picture can be employed as distinguishing features for image matching and retrieval, allowing the correlation of similar images with known geolocations.

This has prompted a number of researchers, for example, \citet{Yuan_geolocation_2022} used computer vision techniques to segment distinguished objects and decorative patterns in images using YOLOv3, which is a pre-trained CNN for object detection. The extracted features are used as inputs to a single neural network for geolocation estimation. The author trained a CNN to predict the location of an indoor image by learning the correlations between the image content and the locations of reference images with known locations. \citet{Nam_Revisiting_2017} suggest combining deep image classification with the Im2GPS (image to GPS) baseline approach, which involves matching a query image against a database of geotagged photos and inferring the position from the return set. By applying kernel density estimation to the positions of a query image's nearest neighbors in the reference database, the authors estimate its geographic location.

While these works achieved promising results, they heavily relied on the availability of reference images with known locations. \citet{Glistrup_Urban-Image_2022} investigate the possibility of geolocating images only based on how well items match publicly accessible geographic information system (GIS) data for real-world applications. This eliminates the requirement for massive collections of annotated images or potentially misleading metadata, which some of the aforementioned techniques rely on. While this approach solves the large collections of geo-referenced data problem, its application is limited to outdoor geolocation and the quality and availability of the GIS data.

With human faces in some images, geolocation of such content could be done through a computer vision generative model that can capture the variations in facial appearance across different regions and ethnicities worldwide. People's appearance and clothing, street sign language, the sorts of trees and plants, and physical aspects of the terrain can all provide semantic clues to the geographic location of a given photo~\cite{Choi_Multimodal_2015}. \citet{bessinger2019generative} developed a generative model based on a statistical analysis of facial features to capture these variations in facial appearance to improve the performance of their computer vision model, thereby enhancing the accuracy and effectiveness of such image geolocation. They use techniques such as principal component analysis (PCA) and Gaussian mixture models (GMM) to determine the underlying distribution of facial appearance variations. The model captures both global and local variables, such as skin color, face shape, and facial landmarks. While the generative model can be used to capture variance in media contents, It is worth saying that it can also be used as a means of manipulating the visual cues for falsification purposes which can lead to wrong location estimation \cite{Anastasia_News_2020}

\subsubsection{Multimodal Analysis} 
Multimodal Analysis involves the use of more than one cue, potentially derived from multiple sources and modalities, such as metadata, image context, audio, and videos for multimedia geolocation. Researchers can gain a more thorough view of the geographical environment by integrating multiple modalities and leveraging the capabilities of each mode~\cite{Salem_Learning_2020, Ma_Multi-source_2018,Jiang_Hierarchy-Dependent_2019}. Using multimodal data allows for the analysis of rich and various information sources, resulting in increased geolocation accuracy, robustness, and contextual comprehension. With multimodal data, content geolocation becomes more complex due to the different data type constitution involved in the analysis, yet enables researchers to glean crucial insights and patterns that would be impossible to identify from individual modalities alone. \citet{Piras_Information_2017} emphasized the importance of fused data for multimedia geolocation and demystified the primary information fusion elements needed from a content-based image retrieval system perspective. The authors highlighted different stages of fusion that can be used to realize richer data for geolocation, among which are early fusion by feature weighting, Late fusion by multi-feature spaces, and fusing various relevance feedback approaches. These are explained in greater detail below:

\begin{itemize}
\item{Early Fusion:} 
This is typically defined as the combining of features into a single representation prior to the computation of image similarity. This is a very common approach in the field of image retrieval, and the simplest and most popular solution is based on the concatenation of the feature vectors into a single vector. Other methods rely on the early fusion of various feature spaces, such as color and shape, or texture and color.

\item{Late Fusion:} 
This refers to the combination of different systems' outputs or the combination of similarity rankings, with the outputs and rankings relating to different feature representations. The outputs to combine are frequently weighted to lend more attention to specific descriptors, either by utilizing predefined \textit{a priori} weights or by learning them for a given image content.

\item{Fusing Relevance Feedback:} 
These strategies include the user in the search refinement process with the aim of bridging the gap between the high-level features, i.e., those observed by human beings that identify semantic information associated to the photos, and the low level ones that are employed in the searches.
\end{itemize}

Many articles utilizing fused data have approached multimedia location estimation using various general and custom techniques and workflows, such as image retrieval systems and scientific workflow. From an image retrieval perspective, \citet{Mookdarsanit_LocationEO_2017} noted that visual content and additional textual metadata of non-geotagged images can be utilized to find similar images from a collection of larger scenes. The authors propose a Geo-signature using visual cues inside the photo that can provide hints about its location MapReduce workflow using feature extraction, geosignature creation, indexing, and similarity matching with computer vision. \citet{Ma_Multi-source_2018} addressed the challenge of estimating coordinates of web images using multi-source attribute fusion including image content, and textual descriptions i.e., metadata, and tags. The authors extract these important features and perform information fusion using a hierarchical strategy that entails computer vision techniques, language models, and machine learning algorithms. \citet{Sethi_Large-Scale_2013} utilize the WINGS workflows, encompassing advanced computing techniques, to scrutinize web content. Through these scientific workflows, they were able to conduct an in-depth analysis of both the text and image components of multimedia web posts, both separately and through a multimodal fusion of the data by leveraging existing workflow fragments and adapting them to perform multimedia analysis. \citet{Friedland_Multimodal_2010} in their experiment to estimate the location of a non-geotagged media recording developed a method for multimodal location estimation by combining information from various modalities that include visual, textual, and contextual cues. The authors extract significant features from multimedia data and map them to related geolocations using techniques from computer vision, natural language processing, and machine learning.

In general, systems that consider a multimodal data approach perform significantly better than those that do not \cite{Friedland_Multimodal_2010, Piras_Information_2017}. However, fused data often have some missing features, such as metadata. This is not always available because these contents can be posted online without such information, and the fused data can introduce false results if altered or wrongly fused \cite{Yuan_geolocation_2022, li_validating_2018}.

\subsubsection{Sun Azimuth-Driven Image Geolocation and Integrity Validation }
The sun azimuth, defined as the angle between the sun's position and a reference direction, can help in multimedia localization by providing useful information about the lighting conditions and scene orientation at the time the multimedia was collected~\cite{Kakar_Authenticating_2012, Li_AreYouLying_2017,li_validating_2018}. Researchers can estimate the approximate direction of media capture by studying the sun azimuth for the authentication of contextual information in geolocated photos or videos. This data has been used to identify outdoor locations, estimate capture times, and verify the validity of multimedia material~\cite{kuznetsov2015new}.

\begin{figure}[h!]
\centering
\includegraphics[width=0.45\textwidth]{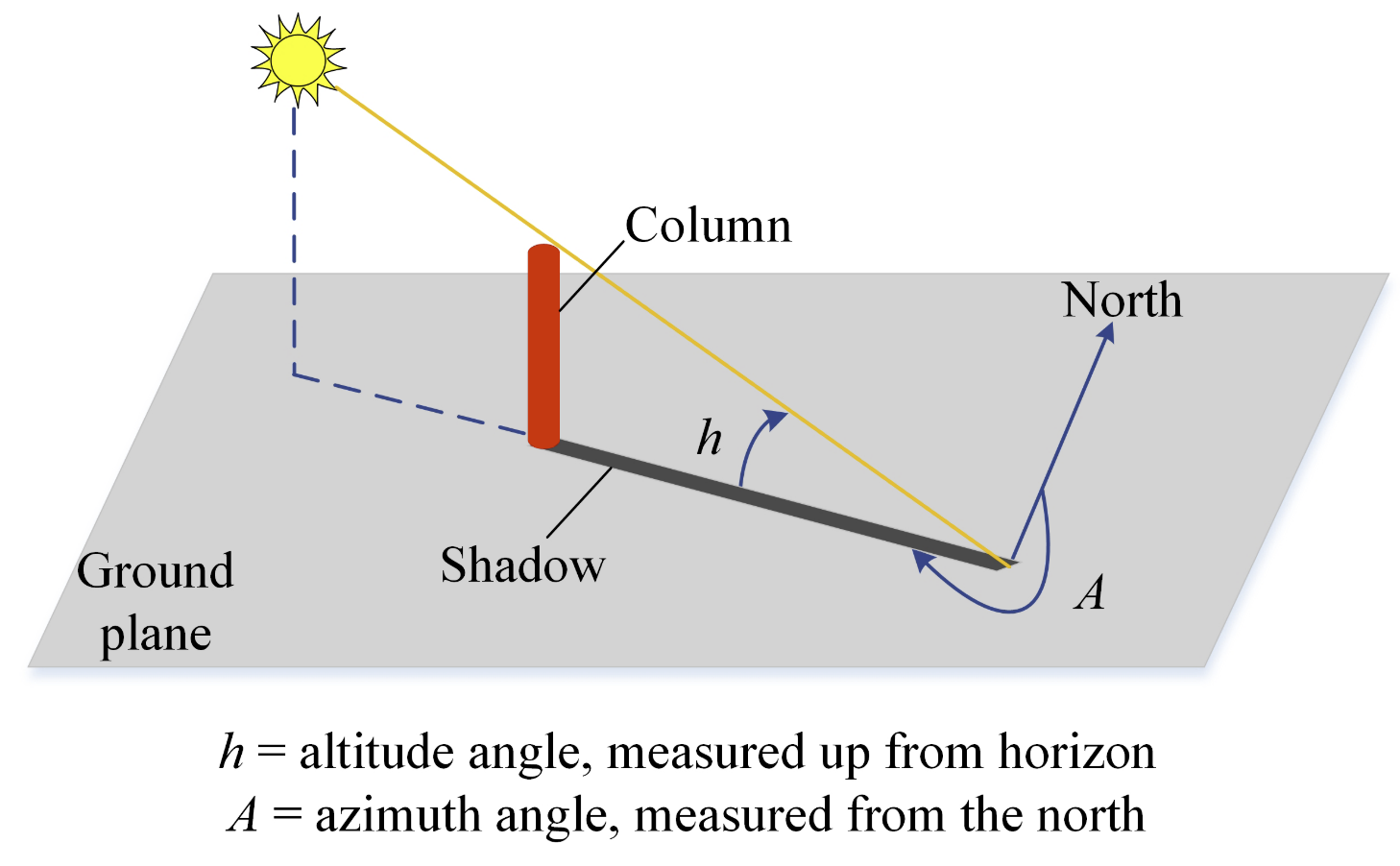}
  \caption{The sun's altitude and azimuth angles illustration~\cite{li_validating_2018}}
  \label{fig:sunAzimuth}
\end{figure}

With the shadows cast by sun reflection on vertical objects, as shown in Figure \ref{fig:sunAzimuth}, the time of capture can be estimated. Sun reflection on objects yields shadows, which can be used to estimate image capture time; influenced by the sun's angle and elevation. Longer shadows occur in the morning or evening, while shorter ones occur around noon. Analyzing shadow length and direction helps in inferring the sun's position. Image capture time was demonstrated by ~\citet{Kakar_Authenticating_2012}, whose focus was on confirming the time of acquisition of a photograph based on its geographical information and a minimum of two-stage shadow detection steps. Inspired by how one of the ancient methods for time-of-day estimation was based on the use of shadows, they developed techniques based on projective geometry to determine the position of the sun in the image by utilizing shadows. To authenticate this information, the authors propose a two-stage approach. The first stage entails determining the sun's azimuthal direction or position based on geolocation data, which is then utilized to determine the time of day when the image was captured. The second stage is cross-referencing the estimated time of capture with the metadata provided by the camera. Similarly, \citet{Li_AreYouLying_2017,li_validating_2018} in their experiment, relied on vanishing points associated with two or more shadows to explore the challenges associated with verifying the authenticity and context of outdoor images through sun position inference. The authors developed a novel approach with astronomical algorithms that uses only one shadow, captures time, and geotags in an image to infer the sun's position. The geolocation is valid only when the inferred and calculated sun positions are consistent. Integrating the sun azimuth with other geolocation signals, such as landmarks or contextual information, might improve the accuracy of multimedia geolocation algorithms.

\begin{figure*}[h!]
\centering
\includegraphics[width=0.8\textwidth]{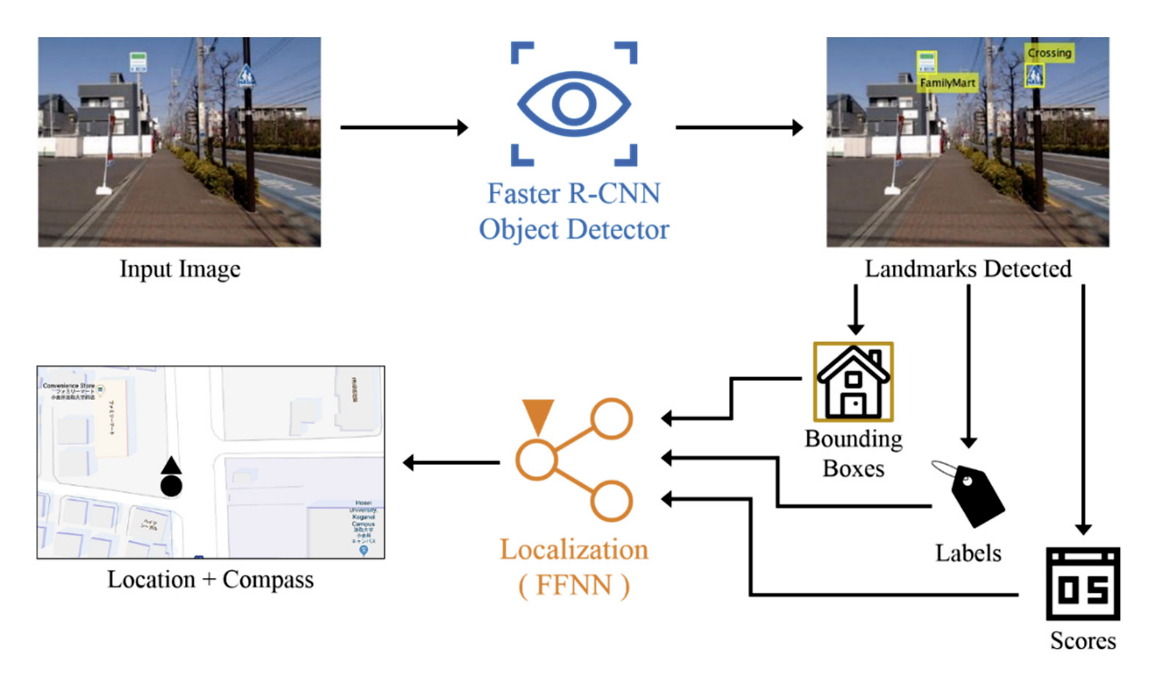}
  \caption{Faster Regional-Convolutional Neural Network (Faster R-CNN)-based localization technique system flows~\cite{Nilwong_DeepLearning_2019}}
  \label{fig:SysFlow_Faster_R-CNN}
\end{figure*}

\subsubsection{Landmark and Skyline Recognition}
It is commonly accepted that humans and other animals utilize landmarks to detect their location in the world and develop routes to their objectives~\cite{Nilwong_DeepLearning_2019}. Computer vision algorithms can detect landmarks, skylines, or notable visual elements in multimedia data. The image's geolocation may be identified by matching these distinguished attributes and characteristics with a database of recognized landmarks, building structures, and visual patterns \cite{Ivanov_Comparative-Study_2013, Zhu_Large-Scale_2022}. There is a trend of employing enormous amounts of data for large-scale vision-based position estimates~\cite{Liangzhi_A-deep-learning-semantic_2021, Nilwong_DeepLearning_2019}. Related approaches, i.e., large-scale picture database indexing, and geo-visual clustering are common to well-defined multimedia geolocation scenarios such as landmark and skyline recognition~\cite{Choi_Multimodal_2015}. \citet{Salem_Learning_2020} demonstrated that learning the feature mapping between ground-level and overhead picture viewpoints enables image localization in places where there are no nearby ground-level images. The authors developed a method for learning and representing the dynamic map of visual appearance that occurs due to changes in lighting conditions, seasons, or other factors over time with machine learning techniques. This was used to extract relevant visual features and to build a model that represents the changes in appearance over time.

\begin{figure*}[h!]
\centering
\includegraphics[width=0.9\textwidth]{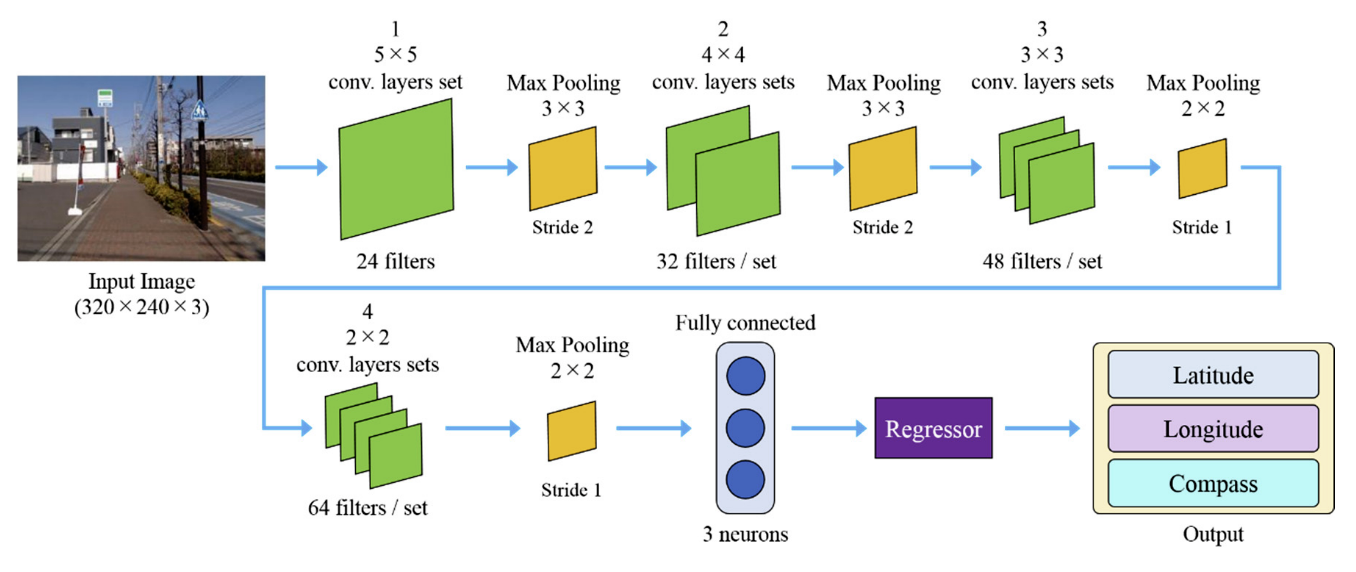}
  \caption{Convolutional Neural Network (CNN)-based localization structure~\cite{Nilwong_DeepLearning_2019}}
  \label{fig:CNN_based_localization}
\end{figure*}

\citet{Zhang_Landmark_2018} explored whether the spatial correlation between the landmark visual and text content contributes to the advancement of multimedia. They begin by introducing a feature refinement network, which learns discriminative and robust visual representations by combining global and local characteristics. Second, to improve retrieval performance, they use a multimodal classifier learning architecture that includes visual and textual inputs. Their experiments on real-world datasets demonstrate the superiority of the proposed approach as compared to existing methods Similarly, \citet{Nilwong_DeepLearning_2019} proposed two methods for outdoor localization based on deep learning and landmark identification. The first method is based on landmark detection in the collected image using the Faster Regional-Convolutional Neural Network (Faster R-CNN). Then, using the observed landmarks, a feedforward neural network (FFNN) is trained to estimate location coordinates and compass orientation, as shown in Figure \ref{fig:SysFlow_Faster_R-CNN}. The second method uses a CNN architecture, as seen in Figure \ref{fig:CNN_based_localization}. The two results both consist of the latitude and longitude angles and magnetic compass orientation. The Faster R-CNN-based method outperformed the CNN-based method at latitude and longitude estimation but suffered a decline in compass accuracy.

\begin{figure*}[h!]
\centering
\includegraphics[width=\textwidth]{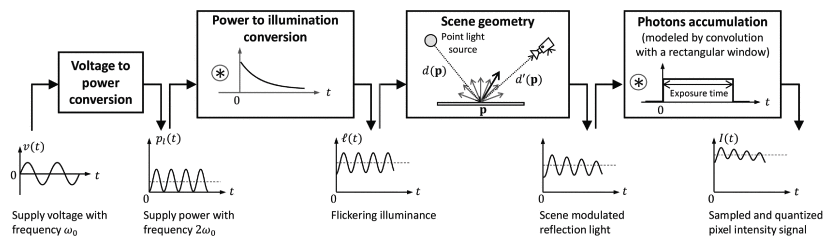}
  \caption{An illustration of the ENF embedding process~\cite{Choi_Invisible_2022}}
  \label{fig:ENF-Embedding}
\end{figure*}

\subsubsection{Latent Inherent Signals and Frequencies Extraction}
It is surprising to see that there is more to intended feature capturing with devices that have many sensors for enhancement. Latent inherent signals and frequencies are sometimes saved along with multimedia files, from which their analysis can reveal cues related to their geolocation. This argument is supported by \citet{Choi_Invisible_2022} extracted invisible Geolocation cues, particularly Electric Network Frequency (ENF) traces, to determine whether it is at 50 or 60 Hz from a single image. The ENF embedding process, as illustrated in Figure~\ref{fig:ENF-Embedding}, shows that its magnitude can be determined. The authors approach this by first quantitatively evaluating the influence of the ENF embedding steps, such as electricity-to-light conversion, radiation dilution due to scene geometry, and image sensing. The verified parametric models of the physical embedding process are then incorporated into a suggested entropy minimization to create a two-level ENF presence classification assessment for region-of-capturing localization. This demonstrates that the basic properties of an image-embedded ENF trace may be correctly detected, and this unique forensic capacity of recovering invisible traces can aid in narrowing down an image's capturing geographical location.


\subsubsection{Regional Distinct Interior Decor Recognition}
This is a computer vision approach used in multimedia geolocation by identifying unique patterns and attributes of indoor settings such as electrical socket types, furniture style and design. By examining these distinguishing aspects in multimedia, such as photographs or videos, it is feasible to deduce the likely place where the media was taken. Image, optical, radio, magnetic, Radio Frequency Identification (RFID), and acoustic sensors are capable of revealing useful information for indoor localization~\cite{Choi_Multimodal_2015, Yuan_geolocation_2022}. WiFi-based indoor positioning capitalizes on the ubiquity of wireless access points (AP) and WiFi-compatible smartphones by estimating the device location based on the signal strength of nearby WiFi beacons.

Several articles addressed the problem of indoor image geolocation, which is a challenging task due to the absence of GPS and the presence of occlusions and clutter in indoor scenes using off-the-shelf image-matching algorithms, such as color histograms, wavelet decomposition, and shape matching~\cite{Stylianou_TraffickCam_2017}. \citet{Yuan_geolocation_2022} use computer vision techniques to segregate objects and ornamental patterns in photos using YOLOv3, a pre-trained CNN for object detection. By understanding the relationships between the image content and the locations of reference photos with known locations, the author trained a CNN to estimate the location of an interior image with promising accuracy. This shows the capabilities of computer vision techniques in indoor geolocation using distinct contextual cues in multimedia.

\subsubsection{Satellite and Overhead Imagery Geolocation}
Other modalities, such as satellite imagery and bird's-eye view imagery, have been considered to further estimate the geolocation of the user photo. Aerial image databases, e.g., Google Earth and Microsoft Bird's Eye View, can be used to learn about geographical representation of the earth~\cite{Choi_Multimodal_2015}. Satellite and overhead imagery can be used to detect the geolocation of multimedia~\cite{Liangzhi_A-deep-learning-semantic_2021, Matasci_Deep_2021}. This method uses the rich spatial information obtained by these imaging sources to extract useful features and match them with known geospatial data \cite{Salem_Learning_2020, Lefèvre_Toward_2017}. The process often involves geolocating pixels which leverage current satellite orbit and attitude parameters with algorithms that use such parameters to determine the latitude and longitude of each pixel~\cite{Ilčev_Global_2019}, but the target of interest typically often takes up a small portion of the image's pixels in a satellite video~\cite{Shao_Tracking_2019, Shao_Can-We-Track_2019, Shao_HRSiam_2021}. A better approach requires preliminary preprocessing e.g., image registration which involves aligning and superimposing two or more pictures from diverse sources, such as photographs or sensors, to produce a composite image in a shared coordinate system. More extensive factors, such as viewing direction, can also be approximated by taking into account geometric relationships between ground-level image pairs, ground-level images, and satellite data, or by aligning images to a 3D scene~\cite{Choi_Multimodal_2015}.

Overhead imagery is readily accessible and offers dense coverage at progressively high resolutions. However, fine-grained features are sometimes difficult to define from the top down. Geotagged ground-level photos are sparsely scattered yet capture high-resolution, semantically rich information. Combining and matching overhead and street-level photographs is difficult using local features such as the output of the SIFT due to the widely variable viewpoint, scale, illumination, sensor modality, and acquisition time, but with advanced computer vision-aided workflows, it is feasible. When supplemental ground-level photos are available, overhead image segmentation can result in considerable accuracy increases~\cite{workman_revisiting_2022}. \citet{Lefèvre_Toward_2017} addressed the difficulty of combining and interpreting satellite images and street-level imagery to get a holistic understanding of scenes. The authors developed a framework that combines deep learning techniques and computer vision algorithms to analyze scenes and extract features from satellite to street level to establish correspondences between different views. This allows the integration of information from both sources for more accurate location estimation. Similarly, \citet{Liangzhi_A-deep-learning-semantic_2021} approach this problem in three steps; 1)  construct a deep learning model to learn semantic representations of images, .2) extract semantic features from the reference and target images using the trained deep learning model, and .3) propose a template matching system that examines semantic properties in order to uncover correspondences between reference and target images, allowing for accurate registration. This provides faster remote sensing picture analysis and pipeline processing, while also allowing for new paths in learning-based registration.

\citet{Choi_Multimodal_2015} relied on the availability of visible structures from both views to align the two planes in such a way that the alignment error is minimized. By changing the field of view (FOV) on the co-located satellite picture, the authors retrieved the FOV from the image to replicate a ground-level view in a certain viewing direction. The horizon of the image is then identified, and a ground plane region is chosen for matching. \citet{workman_revisiting_2022} used a geometry-aware attention mechanism that takes the geographical link between pixels in a ground-level picture and a geographic location into account for segmentation in remote sensing. This can be seen as cross-view geolocation, where ground-level image geolocation estimation is made possible through overhead imagery reference. In addition to this, 13 standard trackers used in traditional visual tracking were thoroughly evaluated by~\citet{Shao_Tracking_2019}. According to their experimental findings, the majority of cutting-edge tracking algorithms rely primarily on brightness, color, or convolutional features, and are therefore unable to follow satellite video objects because these representation attributes are insufficient. The authors suggest a velocity correlation filter approach to get over this problem by building a specific kernel correlation filter for the satellite video target tracking using both a velocity feature and an inertia mechanism.

As established in this section, examination of the various visual patterns and geographical clues present in the aerial photographs increases the feasibility of identifying the location of where the multimedia content was collected precisely. Satellite and aerial imaging are critical in a variety of applications, including surveillance, environmental monitoring, disaster response, image feature matching, image mosaicking, 3D reconstruction, and digital investigations allowing the identification and analysis of locations based on visual evidence. For instance, \citet{Roca_Novel_2016} experiment of an aerial 3D mapping system based on 2D laser scanners for determining point coordinates in outdoor scenes.

\section{Multimedia Geolocation Use-cases}
\label{sec:usecases}
A wide range of use cases emerges with multimedia geolocation, each leveraging the ability of computer vision to address key difficulties. These applications go well beyond standard mapping and image retrieval scenarios~\cite{Sharifzadeh_Vein-Pattern_2014, Roca_Novel_2016}. From assisting law enforcement in the fight against human trafficking to improving navigation systems, multimedia geolocation is emerging as a vital tool with far-reaching ramifications. This part digs into the vast landscape of use cases, investigating how computer vision techniques are used to pinpoint places and unravel the intricate tapestry of their real-world impact.

\subsection{Organization and Retrieval of Media}
With the continuous and exponential increase in the rate at which media are generated, uploaded, and transferred from one device or platform to another, effective organization and retrieval of media including corresponding geolocations, are essential for efficient data management, analysis, validation, and personalized experience in various services and industries 
including e-commerce, marketing, tourism, publishing, medicine, criminology, sports, culture, and entertainment\cite{Zhang_Landmark_2018, Zhu_Large-Scale_2022, lallie_geotract_2013, Choi_Multimodal_2015, Friedland_Multimodal_2010, Keerthi_RIPA-2018}. By implementing advanced techniques such as content-based indexing, semantic tagging, and similarity search algorithms, multimedia content can be compared, investigated, and navigated \cite{Stylianou_Indexing-open_2015, Ivasic-Kos_Application_2022,moreira_image_2018, Zhang_DeepDive_2017}. Search engines such as Google, Yahoo, Bing, Getty, Flickr, Shutterstock, and Yandex are continuously dedicating more resources to improving the geolocation of multimedia data for a more personalized user experience~\cite{Zerdoumi_A-new-spatial_2022}. \citet{Friedland_Multimodal_2010} highlighted that a rising number of websites now offer public APIs for organized access to their material, and many of these include geolocation features. Flickr, YouTube, and Twitter all allow you to search for results that originated in a specific location.

\citet{Ivasic-Kos_Application_2022} focused on the use of digital images and the image retrieval paradigm in a variety of applications, such as locating relevant images based on their visual similarity. Through ensemble learning that combines feature refinements, such as automatically geotagged images with low-rank matrix recovery and multimodal classifier learning,~\citet{Zhang_Landmark_2018} improved landmark image retrieval and recognition accuracy and robustness. It is important that any solution aimed at addressing the organization and retrieval of multimedia issues effectively must rely on many image representations that must be easily fused to adapt to the subjectivity of image semantics\cite{Piras_Information_2017}.

\subsection{Map and Navigation Systems}
Multimedia geolocation is critical for improving maps and navigation systems. These systems can provide consumers with a more immersive and detailed experience by combining multimedia content with geographical information. In cartography, Which relies on spatial data and techniques to represent and understand geographic information, geolocated multimedia can be integrated with cartographic data and visualization techniques to enhance the understanding and interpretation of geospatial information~\cite{Zerdoumi_A-new-spatial_2022}. The availability of a huge number of geotagged ground-level images from social media and camera-equipped vehicles has enabled the creation of various novel mapping systems~\cite {workman_revisiting_2022}. Geolocated multimedia allows users to navigate with greater precision and context by accurately placing landmarks, sites of interest, and other relevant information on the map. It also aids in the visualization of real-world scenes and environments, allowing users to better understand their surroundings. With multimedia geolocation, dynamic and interactive maps incorporating street-level photography and 3D models facilitate users to have a more interesting and informative navigation experience~\cite{Choi_Multimodal_2015}.

This is currently a feature of Google Maps among other 2D web mapping applications including Yahoo  Maps,  Bing  Maps,  OpenStreetMaps,  ArcGIS  Online,  and  3D  virtual globes e.g.,  Google  Earth, Bing Maps 3D, ArcGIS Explorer, and Marble~\cite{Zerdoumi_A-new-spatial_2022}. This demonstrates the confluence of ground-level images and remote sensing capabilities as discussed by \citet{Lefèvre_Toward_2017}. The authors shared their thoughts on Mapillary; a crowdsourcing endeavor that intends to distribute geotagged photographs to provide an immersive street-level perspective of the world. Its 100 million images are used by computer vision algorithms to detect and distinguish objects, perform 3D modeling, and comprehend text/signs. In addition to crowdsourced geotagged photographs, Google Street View, Bing Maps Street-side, and Apple Maps provide ground-level views obtained through planned mobile mapping efforts on a global scale. Similar initiatives have also been seen in China (Tencent Maps), South Korea (Daum Maps), France (Mappy), Switzerland (GlobalVision Video StreetView), and European and US cities (CycloMedia)~\cite{Choi_Multimodal_2015}.

\subsection{Law Enforcement and Digital Investigation}
The task of geolocation has become a standard component of the toolset for investigative journalists and academics who deal with image and video files to verify the validity of the content~\cite{Glistrup_Urban-Image_2022, Neal_Global_2019, lallie_geotract_2013}. Law enforcement organizations have a serious difficulty as the amount of illicit media shared on peer-to-peer (P2P) networks and other similar platforms grows~\cite{Claudia_Live-forensics_2016}. Geolocation information is not only critical in criminal investigations, but it is also becoming increasingly significant in digital forensics, since it enables the logical synthesis of digital evidence that is frequently scattered across multiple mobile assets~\cite{Al-Kuwari_Probabilistic-2010}. Law enforcement agencies analyze and interpret multimedia data to gather evidence, establish timelines, and geolocate individuals or objects of interest in digital investigations. Computer vision-based geolocation has the potential to improve investigators' capabilities in geolocating multimedia content, reconstructing events, and ultimately supporting efforts to combat various criminal activities, including human trafficking, by integrating computer vision into digital investigation methodologies.

\citet{Yokota_Visual-based_2020} identified the need to replace the geolocalizing element in the traditional visual-based geolocalization tool with machine learning techniques to assist forensic investigators in determining the location associated with evidential images. The authors developed a visual-based geolocalization framework to enable reliable and accurate geolocation information to support investigations. The potential of Geolocation cues along with other information found in the metadata of multimedia contents was highlighted by \citet{moreira_image_2018}. The author had a specific focus on media provenance analysis to gain insights into the origin and history of digital images, which is a frequently encountered issue in digital investigation. \citet{Claudia_Live-forensics_2016} developed an innovative live forensics approach called iCOP (Intelligence in Cyber and Open Source P2P) to uncover previously unknown criminal media on P2P networks and ~\citet{lallie_geotract_2013} developed the GeoTract framework that employs R-tree spatial indexing techniques as well as parallel processing to extract geotag pairs and IP addresses and maps them using Google Maps API. The authors hope to improve the efficiency and effectiveness of detecting and capturing unlawful content geographically by employing real-time monitoring and analytical tools.

\section{Multimedia Forensics in Human Trafficking Investigations}
\label{sec:forensicsHT}
Combating human trafficking is riddled with issues that necessitate not only research, but also a thorough understanding of the digital fingerprints left by these terrible actions~\cite{Horan_cyber_2021}. As traffickers leverage various digital mediums, multimedia content analysis takes center stage in uncovering clandestine networks and assisting law enforcement. The methods used in multimedia forensics are as varied as the difficulties they solve, ranging from image provenance analysis to the detection of temporal metadata modification~\cite{Kim-Kwang_Forensic-Visualization_2017}. These are further discussed in the following sections. 

\subsection{Multimedia Evidence Prevalence in Human Trafficking Investigations} 
In-depth context and insights are provided to law enforcement and related agencies through multimedia evidence, which plays a vital role in human trafficking investigations. Digital evidence identification and processing are also crucial to many child-focused investigations~\cite{Anda_Assessing_2020}. Modern-day traffickers frequently employ digital platforms and communication methods to carry out their illegal acts, leaving a trail of multimedia content including text descriptions, images, videos, and audio recordings in their wake. The patterns, connections, and places connected to human trafficking organizations can be found using these digital artifacts, which can be a priceless source of information for investigators. The abundance of multimedia evidence highlights the need for cutting-edge methods in computer vision to evaluate, interpret, and geolocate such content, allowing law enforcement to thoroughly comprehend the scale of human trafficking networks, identify victims, and prosecute offenders~\cite{Swaminathan_Predict_2022}.

\citet{Sethi_Large-Scale_2013} in their work towards HT detection of minors analyzed fused multimedia content from the web using a scientific workflow. Similarly, \citet{Paul_Multi-modal_2020} utilized the fusion of text features generated with text vectorization technique and image features in their deep learning modelling. \citet{Sukumar_Open-research_2015} collected millions of photographs from the web and applied cutting-edge learning techniques to automatically provide a conceptual description of each image. \citet{kamath20212021} approach this problem by narrowing it down to hotel identification. The authors believed that since victims are frequently photographed in hotel rooms, hotel recognition is a crucial duty for human trafficking investigations. \citet{Mattmann_Scalable_2017} noted that people who are involved in criminal activities such as HT on the dark web often employ multimedia content as a way to avoid being found by bulk analysis and traditional online searches. The authors gathered a sizable video collection from websites that specialize in HT to perform content detection and analysis using a Hadoop-based algorithm after several trials with Pooled Time Series.

In addition, \citet{Ramchandani_Unmasking_2021} in their work to unmask HT Risk in commercial sex supply chain using machine learning utilized unstructured deep web data to characterize the scale of trafficking recruitment and sales risk. The authors developed a geographical network picture of the supply chains for commercial sex, from hiring to selling, using this data in conjunction with a cutting-edge machine learning framework. \citet{Varastehpour_Vein_2019} used cutting-edge methodologies and computational tools, such as optical-based vein uncovering and artificial neural networks, to recognize vein patterns in evidence photos for forensics identification.

Investigations into human trafficking can be made more thorough, fast, and effective in preventing this serious violation of human rights by utilizing more multimedia evidence.

\subsection{Digital Forensics of Multimedia Evidence in Combating Human Trafficking} 
The discipline of collecting and organizing information found on an electronic device for investigation purposes is known as digital forensics. It is critical to understand both the technologies and the strategies and frameworks used by investigators in this discipline~\cite{Alruwaili_CustodyBlock_2021}. There are four types of digital forensics: host forensics, mobile forensics, network forensics, and cloud forensics. Each of these four regions gives various types of information to investigators, with little overlap in multimedia evidence~\cite{Horan_cyber_2021}.

The majority of crimes are supported by digital evidence, and nearly all criminal investigations include analysis of this data. Due to the sharp rise in both the number of cases requiring digital forensics analysis and the amount of data that must be processed for each case (as a result of increases in the number of pertinent devices and their capacities), the demand for digital forensics investigation has skyrocketed~\cite{Anda_Assessing_2020, salamatcitesranger, Granizo_detection_2020} therefore, the digital forensics team cannot avoid implementing more robust and resilient evidence-handling techniques~\cite{Alruwaili_CustodyBlock_2021} just as \citet{GALBRAITH2020301009} noted the growing need for the development of quantitative statistical methodologies in digital forensics. Even if we set aside the multiple difficulties in gathering and analyzing digital data for the time being, there are many complex problems with how the various stakeholders share the evidence~\cite{Casino_Sok_2022}.

While there are a number of digital forensics procedures, models, or frameworks, a good number of these have some elemental units relating to four major processes which are ``identifying, preserving, analyzing, and presenting''~\cite{Kim-Kwang_Forensic-Visualization_2017, bohme2016media}. These processes require various tools depending on the nature and format of the evidence and computer vision is more useful for contextual media forensics which is key to combatting HT~\cite{bohme2016media, Srivastava_CamForensics_2017}. For instance, \citet{Sharifzadeh_Vein-Pattern_2014, Varastehpour_Vein-Pattern_2019, Keivanmarz_Vein-Pattern_2020} utilized vein pattern as a biometric attribute for forensic identification in crime investigation using computer vision techniques.

Can we create a multimedia tool that automatically detects illicit human trafficking regardless of age, gender, color, or culture? This is a crucial question by ~\citet{Karnik_A-new_2022} aimed at assisting the investigation agencies. Experts in digital forensics use specialized tools and procedures to verify the origin and integrity of multimedia files, recover hidden or erased data, and pinpoint the time and location of creation.

Comprehensive data is crucial in the fight against human trafficking. The data and its analysis would give the organizations involved in counter-trafficking a better understanding of the patterns, scope, and tactics of trafficking activities~\cite{Upadhayay2021CombatingHT}. Examining both text and visual data is necessary for this challenging endeavor~\cite{Sethi_Large-Scale_2013} but due to the shortcomings of current forensic technologies, there is a considerable backlog in gathering, accessing, organizing, integrating, and analyzing enormous amounts of data from a range of data sources as earlier established.

In response to some of these issues, \citet{Varastehpour_Vein_2019} noted that researchers, law enforcement agencies, and other concerned bodies can combat this expanding crime by developing a well-established identification system based on non-facial skin scans that aid in the identification of pedophiles. Contrary to this, \citet{chhoriya2019automated} used a Haar feature-based cascade classifier to develop an automated facial recognition system for criminal identification. \citet{Sukumar_Open-research_2015} proposed a method that can automatically characterize or tag image information in order to address the prevalent problem of image triaging by evaluating the art of the possible with machine learning. This also piqued the interest of \citet{Sukumar_Open_2015}, who scrapped millions of images from the web and evaluated the art-of-the-possible with machine learning to provide a solution to the common image triaging problem by proposing a system that can automatically describe or tag image content.

\citet{Vajiac_TrafficVis_2023} developed a tool called TRAFFICVIS for cluster-level HT detection and labeling. Specifically, the authors extend state-of-the-art text clustering algorithms and visualize the outcomes by adding shared metadata as an indicator of related and potentially suspect activity. \citet{Stylianou_TraffickCam_2017} created a platform for crowdsourcing, TraffickCam, enabling the public to submit anonymously hotel room pictures for feature extraction, representation, and indexing from which law enforcement organizations can extract masked images of sex trafficking victims for appropriate matching. Through the examination of suspicious clusters of linked adverts, law enforcement, and domain specialists can find human trafficking (HT) in online escort websites. Contrary to this, ~\citet{dasgupta2022audio} focused their attention on developing an innovative audio analytics-based human trafficking detection framework using 1-D CNN for autonomous vehicles. According to the authors, violent language and aggressive behavior can be signs of illegal activity like human trafficking as well as violent crime. A wide network of human traffickers may be located and eventually, their routes and network may be revealed if such behavioral patterns or tones are identified.

In order to aid in the investigation of cases and the speedy identification of victims,~\citet{Gowda_An-Approach_2017} developed a fast, efficient and easily deployable framework automatically performing image forensics just as \citet{Vinavatani_AI_2022} developed a system that will aid in the identification of missing people through the use of face recognition, machine learning, deep learning, and artificial intelligence. Similarly, \citet{Li_Detection_2018} describes an unsupervised and scalable template-matching technique for analyzing and recognizing sophisticated organizations that operate on adult service websites. The authors used the advertisement content to discover signature patterns in text that are indicative of organized activities and organizational structure. Take Twitter as an example of an online environment that traffickers often use for advertisement. \citet{Granizo_detection_2020} developed a system that recognizes tweets that could promote these illegal services and exploit minors using Natural language processing. \citet{Stylianou_Indexing-open_2015} proposed a platform that incorporates an open-source information database available on the internet, a crowdsourcing strategy to gather hotel photos for quick identification, and location matching to combat HT. \citet{Kejriwal_technology_2018} instead developed a Domain-Specific Insight Graphs which is a complex ensemble of various adaptive technologies to help users perform investigative search. \citet{Parolin_3M-Transformers_2021} in their work aimed at providing enough structured data from unstructured data for effective application of computational methods explored 3M-Transformers (Multilingual, Multi-label, Multi-task) for event coding from multilingual, domain-speciﬁc corpora. The integration of cutting-edge technologies such as computer vision into digital forensics has ushered in a new era in the fight against human trafficking.

\subsubsection{Kinship}
By comparing two people's appearances, kinship verification attempts to ascertain whether they are biologically related. There are numerous uses for kinship verification and family classification, including tracking missing children, creating family trees, managing multimedia, and tagging people on social media~\cite{Robinson_The_2021}. The ability to destroy trafficking networks, find victims, and safeguard the safety and well-being of those affected by HT can be improved by kinship analysis's integration with computer vision for multimodal geolocation~\cite{Serraoui_Knowledge-based_2022, Wang_Cross-Generation_2019, Goyal_Family_2022, Haibin_Multi-scale_2021, Yadav_A-Feature-Averaging_2019, Liu_Age-Invariant_2022}.

When using computer vision for multimedia geolocation, kinship analysis is very important in the battle against human trafficking. Families are frequently used by human traffickers to influence and coerce their victims and due to the stark disparities between family members' faces, the largest barrier to kinship verification is that extracted features may not have a strong representational capacity~\cite{Chen_Semi-coupled_2021}. Kinship analysis can help law enforcement and investigation teams locate these connections and disentangle the intricate webs of trafficking networks~\cite{Chandaliya_Conditional_2019, Robinson_Visual_2018, Robinson_To-Recognize_2018}.

For instance, \citet{Goyal_Family_2022} in their work on family classification proposed a novel weighted nearest member metric learning technique based on simultaneously learning numerous measures to guarantee the separation of members of negative families and the compactness of members of positive families. \citet{Goyal_Patch-Based_2021} proposed a novel method that compares entire facial photos using the Global-based Dual-Tree Complex Wavelet Transform. The authors noted that through the use of patch-based methods, kinship recognition can be achieved with improved accuracy and efficiency. Similar to this, ~\citet{abbas2022kinship} built a model of a deep relational network that leverages the age modification of two facial photos as a preprocessing step using a Siamese network with two CNN before comparing them to draw out family ties with the aim of locating missing children and stop human trafficking. \citet{Haibin_Multi-scale_2021} used the convolution operation's basic idea to extract several scales of features from a face image, which depicts face features from various angles. Similarly, to lessen the discrepancies between facial photos, \citet{Chen_Semi-coupled_2021} suggested a semi-coupled synthesis and analysis dictionary pair learning (SSADL) method. The authors believe that the heterogeneous facial images of the parent and child can be transformed into coding coefficients of the same subspace using SSADL, which learns jointly two view-specific synthesis-analysis dictionary pairs as well as a mapping matrix from the parent and child training data. The kinship verification task can then be carried out using the coding coefficients.

From a multimedia geolocation perspective, Kinship analysis can help connect people across multiple regions and eras. Computer vision approaches can identify familial relationships even when victims are separated or transported across different regions by examining visual and biometric cues. With the use of this information, investigators can put the wider picture of human trafficking operations together and identify the major players and networks that are engaged.

\begin{figure*}[h!]
\centering
\includegraphics[width=\textwidth]{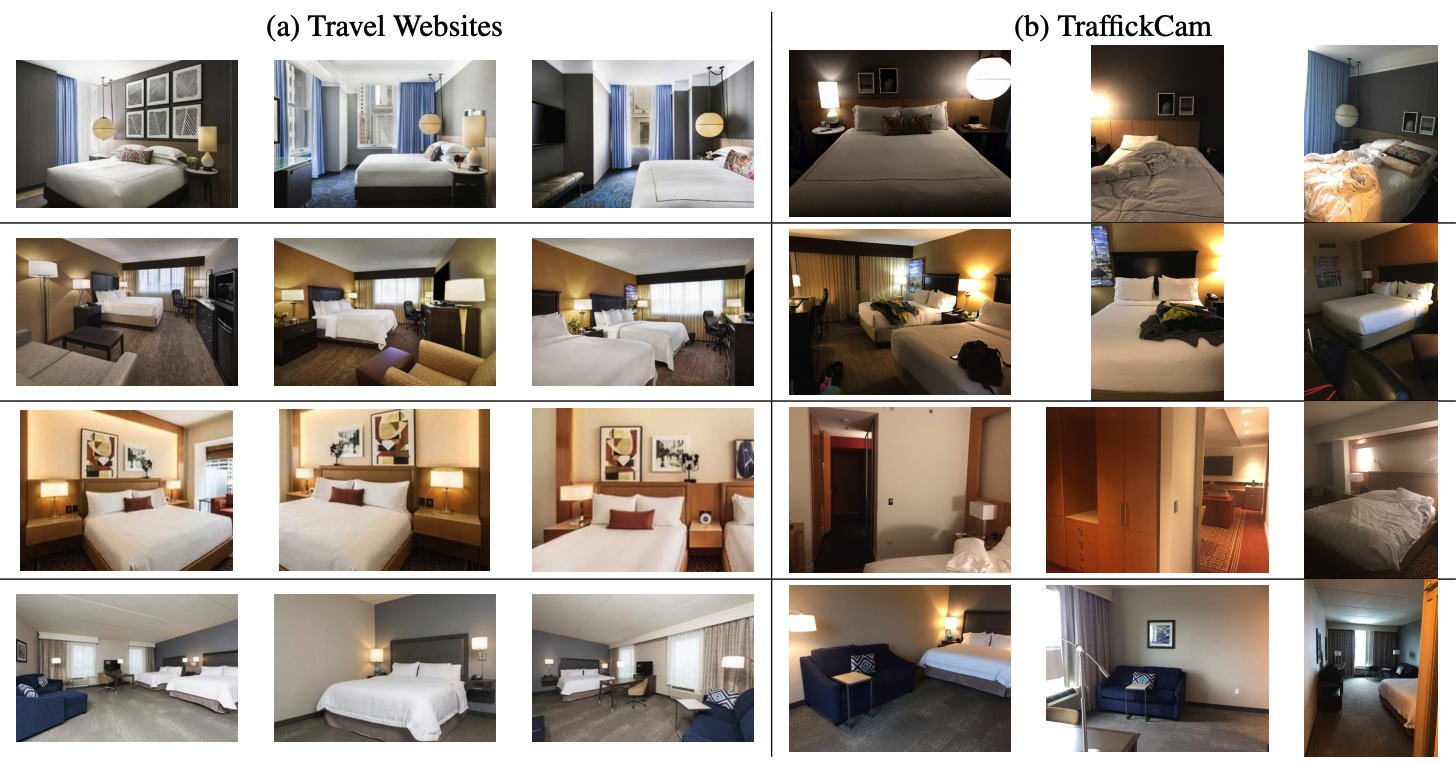}
  \caption{Samples of images in the Hotel 50K dataset and sources (a) travel website and (b) TraffickCam App Database~\cite{Stylianou_Hotels-50K_2019}}
  \label{fig:Hotel-50K}
\end{figure*}

\subsubsection{Hotel Room Identification}
With the continuous utilization of multimedia content in trafficking advertisements on the open and dark web, it has been noticed that the majority of these contents relating to trafficking adverts with the aim of connecting clients together are taken in different hotels. Given that victims of human trafficking are frequently photographed in hotel rooms, finding these hotels is essential to trafficking investigations since it allows investigators to locate past, present, and potential victims who may have been trafficked to the same locations~\cite{kamath20212021}. The Hotels-50K dataset introduced by \citet{Stylianou_Hotels-50K_2019} consists of over 1 million images from 50,000 different hotels around the world. These images are sourced from both travel websites and the TraffickCam mobile application~\cite{Stylianou_TraffickCam_2017}, which allows regular travelers to upload images of their hotel rooms to aid in the fight against human trafficking. Photographs from TraffickCam look more like images from trafficking investigations than images from travel websites, as seen in Figure~\ref{fig:Hotel-50K}. There are 1,027,871 images in the training dataset from 50,000 hotels and 92 major hotel companies. Of the 50,000 hotels, 13,900 have user-submitted photographs from the TraffickCam application (the training set has a total of 55,061 TraffickCam images). The test dataset contains 17,954 TraffickCam photos from 5,000 hotels.

\subsection{Impact of Variational Factors in Recognition and Location Algorithms} 
The digital examination of trafficking victims might be strongly impacted by variable factors, such as aging and changes in face appearance. These elements make it more challenging to identify and find victims over time in certain circumstances~\cite{Anda_Evaluating_2018}. Due to stress, trauma, or the passage of time, those who have been trafficked may experience physical changes that change their facial characteristics and overall look~\cite{Koudelová_Modelling_2014}. Due to this, it may be challenging for law enforcement and investigators to compare current multimedia content, such as pictures or videos, with old documents or databases. The ability to observe these changes over time can help detectives build more precise profiles and increase the likelihood of finding victims~\cite{vyawahare2022synthetic}. This has dramatically improved with the advent of AI and its cutting-edge branches, such as computer vision, machine learning, and deep learning~\cite{Anda_Improving-Borderline_2019, 10.1145/3407023.3407068, Mohamadi_Human_2021} giving researchers like~\citet{Anda_Evaluating_2018} the opportunity to evaluate the automated facial age estimations techniques for digital forensics.

Considering these factors, various researchers address this issue from different perspectives, such as the inclusion of age estimation algorithms (a valuable tool in the field of multimedia geolocation for locating and identifying people over time \cite{10.1145/3407023.3407068}) and generative artificial intelligence, such as Generative Adversarial Networks (GAN) in their proposed HT detection pipeline. Also with the help of old photos of the missing child, forensic age progression can anticipate the missing child's current appearance, which can be important information in cases involving long-term missing children~\cite{Haibin_Multi-scale_2021}.

With GAN demonstrating remarkable performances in various computer vision tasks including image translation, image generation,  face image synthesis, super-resolution imaging, and many more,~\citet{Wang_Cross-Generation_2019} investigated the possibility of using image translation to create an intermediate domain to reduce differences in age and identity. The authors proposed a cross-generation generative kinship verification technique that combines a sparse discriminative metric loss, a deep architecture, and a towards-young generative model into a single framework.in parallel to this, ~\citet{Chandaliya_Conditional_2019} their work to address face aging and rejuvenation on children's faces, designed a novel architecture based on GAN and Variational AutoEncoder with Perceptual Loss. Subsequently, \citet{Chandaliya_child_2021} created a methodology for child face age progression and regression that creates identity-preserving, photo-realistic face images. The authors use a Self-Attention Block to learn global and long-term dependencies within an internal representation of a child's face and a Multi-scale Patch Discriminator Learning Strategy to train Conditional Generative Adversarial Nets, which increases the stability of the discriminator. Likewise, \citet{Liu_Age-Invariant_2022} in their work on kinship verification, designed a module for learning age-invariant adversarial features that is effective at extracting such features to have a reliable result from their solution system pipeline. Though the efforts in accounting for these variabilities are yet to reach the desired peak, addressing these factors through innovative approaches will be essential for enhancing the accuracy and effectiveness of recognition and location algorithms as technology continues to evolve.

\section{Discussion}
\label{sec:discussion}

Following a thorough analysis of the use of computer vision-based multimedia geolocation approaches for human trafficking, from a digital forensic investigation standpoint, it is prudent to discuss the effect of this technology. This discussion aims to illuminate the areas where computer vision has shown its effectiveness through the research questions that were earlier identified and to identify areas that require further refinement, serving as a guide for future research directions.


\subsection{RQ1: What are the state-of-the-art computer vision techniques used in the context of multimedia geolocation in digital forensics?}
The examination of this research question in the literature uncovers a compelling landscape of innovation and promises. Several notable techniques have emerged as front-runners in this field, offering robust solutions to the multifaceted challenges posed by multimedia geolocation. While some approaches cover the entire planet, the majority operate on a much smaller scale, such as within a single city. Early techniques often treated this as a picture-matching problem, using geotags retrieved from matched database photographs. Recent research has focused on estimating finer-grained and more comprehensive geo-context information, such as viewing direction estimation~\cite{Choi_Multimodal_2015}.

The state-of-the-art computer vision in multimedia geolocation is continuously evolving around deep learning-based approaches, with classification and retrieval methods demonstrating exceptional capabilities. Both of these networks and techniques are adept at feature extraction and can discern intricate patterns within images, making them indispensable tools for geolocation tasks.

With classification techniques, the categorization and organization of multimedia data based on predetermined classes or criteria allows for more effective geolocalization. Deep learning and machine learning algorithms are trained on visual features to categorize multimedia material into different geographical classes. These algorithms recognize patterns associated with certain landmarks, architectural styles, or environmental circumstances. Researchers using the classification approach frequently achieve improved accuracy, fine-grained localization, efficient processing, interpretability, and generalization~\cite{ Claudia_Live-forensics_2016, Yuan_geolocation_2022, Stylianou_TraffickCam_2017, Friedland_Multimodal_2010}.

Search and retrieval techniques enable efficient exploration and retrieval of relevant multimedia content based on specific search queries or similarity metrics. While there are numerous techniques for achieving this, content-based analysis with computer vision has been proven to be popular among researchers due to the high relevancy of resulting searches. To retrieve geotagged multimedia data from a database that closely matches a query image, Content-Based Retrieval solutions use similarity-based search approaches that rely largely on embeddings or vectorization techniques and lazy learning algorithms. This method works well if there are images in the database with fields that significantly overlap the query image fields. The main advantages of this mode are query flexibility, scalability, adaptability, the identification of unexpected locations, visual similarity investigation, and geolocation granularity flexibility~\cite{Nam_Revisiting_2017, Sethi_Large-Scale_2013, Stylianou_Indexing-open_2015, Glistrup_Urban-Image_2022}.

With the level of complexity and difficulty in building generalized models and solutions for multimedia content geolocating, the categorization of the data into indoor and outdoor scenes increases the chances of attaining computer vision technique's promised results~\cite{Yuan_geolocation_2022, Stylianou_Indexing-open_2015}. Separating this problem into indoor scenes and outdoor scenes has shown promising directions toward building robust and domain-specific geolocation systems based on the conglomeration of computer vision techniques. Researchers are effectively leveraging rich data such as GPS-tagged images as well as a wide range of visual cues including but not limited to landmarks, skyline, street signs, satellite views, and sun azimuth angle for outdoor scene geolocation~\cite{Ivanov_Comparative-Study_2013, Zhu_Large-Scale_2022, Kakar_Authenticating_2012, li_validating_2018} while solutions to indoor scene geolocation can be more specific depending on the pattern of the problem as seen in some of the works earlier reviewed where authors used cues such as regional distinct interior decoration recognition and latent inherent signals and frequencies extraction~\cite{Choi_Invisible_2022, Yuan_geolocation_2022}.   

The effective application of computer vision techniques is highly dependent on the multimedia data format which includes images, videos, and text from sources as diverse as social media platforms, surveillance cameras, and satellite imagery~\cite{Glistrup_Urban-Image_2022, Friedland_Multimodal_2010, lallie_geotract_2013}.CNNs and deep learning models have excelled at object recognition and image classification tasks for image-based content analysis, assisting in the identification of trafficking-related features in images such as those on social media. In the case of video content, spatio-temporal analysis techniques allow for the tracking of objects of interest in numerous frames, allowing for the reconstruction of trafficking patterns or the identification of movement patterns.

Optical object recognition coupled with NLP algorithms plays an important part in analyzing text-based multimedia such as media metadata, descriptions, and captions accompanying multimedia data which encode numerous cues for geolocation~\cite{Murdock_2016, Keerthi_RIPA-2018}. With these techniques, researchers can extract useful information embedded in multimedia, recognize keywords, and perform other analysis just as~\citet{Hundman_Always-Lurking2018} did in developing an HT detection system capable of identifying internet advertisements as trafficking-related or not.

\subsection{RQ2: How is computer vision contributing to efforts aimed at countering human trafficking?}
The lack of actionable intelligence regarding trafficking occurrences and operations is a serious impediment in the fight against global human trafficking and as established earlier, it takes great effort from different disciplines and viewpoints to comprehend and counter such criminal networks. Through various key achievements, computer vision techniques will continue to play a pivotal role in the development of efforts to combat human trafficking.

\citet{Yokota_Visual-based_2020} identified the necessity to replace the localizing element in the standard visual-based geolocalization tool with state-of-the-art machine-learning approaches to help forensic investigators establish the location associated with the evidence. These techniques have significantly increased the ability to analyze and extract relevant information from shreds of digital evidence, which are regularly used to document and disseminate trafficking-related actions. Multimedia analysis with computer vision techniques such as object detection, image segmentation, and facial recognition enable the identification of suspects, victims, and repetitive patterns in human trafficking operations. Also, deep learning models have enhanced image and video classification and retrieval, aiding in the automated detection of explicit content or signs of trafficking in multimedia data.

Social media monitoring for illicit acts is also necessary. Human traffickers frequently utilize social media platforms to recruit, advertise, and communicate. Computer vision algorithms are capable of scraping, analyzing, and monitoring massive volumes of social media data, detecting dubious content, and identifying traffickers and their networks~\cite{Hundman_Always-Lurking2018} just as demonstrated by \citet{tong_combating_2017} who developed a multimodal deep model termed the Human Trafficking Deep Network to identify human trafficking advertisements in the Trafficking-10K dataset. This proactive method allows law enforcement to intervene and prevent human trafficking.

Fused data comprising images, videos, text, and audio recordings allows investigators to build a complete picture of trafficking operations, which aids in the dismantling of criminal networks \cite{Piras_Information_2017, Mookdarsanit_LocationEO_2017}. With computer vision techniques, these multimodal data analyses and insight derivations are made easier and often serve as benchmarks for investigative evaluations. This was demonstrated by \citet{Ma_Multi-source_2018} who attempted to solve the difficulty of calculating coordinates of image evidence collected from the web using a hierarchical strategy comprising computer vision techniques.

The automation of processes such as evidence processing, categorization, and data extraction saves investigators time and resources. Computer vision speeds up the investigation process through fast and efficient object of interest recognition, identification, segmentation, and enhancement algorithms, helping law enforcement to respond to human trafficking events more quickly. \citet{Gowda_An-Approach_2017} developed a fast, efficient, and easily deployable framework for automatically performing image forensics just as \citet{Vinavatani_AI_2022} developed a system that will aid in the identification of missing people through the use of face recognition, machine learning, deep learning, and artificial intelligence.

\subsection{RQ3: What are the potential implications of multimedia geolocation for human trafficking investigation and prosecution?}

One of the most important components of human trafficking investigations is pinpointing the place where the multimedia content was made or shared. Multimedia geolocation has far-reaching ramifications for evidence collecting and prosecution in human trafficking cases. Computer vision paired with geolocation data aids in determining the origin of digital evidence. Investigators can geolocate the origins of the content in trafficking cases by analyzing visual indicators in photographs and videos such as landmarks, street signs, or unique geographical features~\cite{ Neal_Global_2019, lallie_geotract_2013}. 

Multimedia geolocation gives a solid framework for prosecuting traffickers. Law enforcement organizations can establish a strong link between suspects and criminal actions by pinpointing the particular location where evidence, such as incriminating images or videos, was made or transmitted~\cite{Claudia_Live-forensics_2016}. This spatial context not only strengthens the evidence's reliability but also aids in correlating victim testimonies and identifying co-conspirators.

Victims of human trafficking frequently provide vital testimony, but their recollections may be hazy due to trauma or fear. Multimedia geolocation is an objective tool for verifying victim statements. Geolocation data may verify the veracity of victims' claims and strengthen the reliability of their testimonies in court when they retell their experiences and describe locales.

Trafficking activities may take place in numerous locations. Multimedia geolocation assists investigators in locating trafficking hubs where victims are detained or exploited. Law enforcement can deconstruct entire trafficking networks by mapping these centers and connecting them to suspects, leading to more complete convictions. Knowing that human trafficking has evolved in the digital age, with criminals employing internet channels for recruiting and advertisement. Multimedia geolocation aids in following traffickers' online actions and determining the geographical origin of online adverts. This was demonstrated by \citet{lallie_geotract_2013}, who developed a GeoTract framework that uses R-tree spatial indexing techniques as well as parallel processing to extract Geo-tag pairs and IP addresses, which are then mapped using Google Maps API. This evidence is critical for prosecuting online trafficking cases.

Finally, geolocation data is also critical for preserving evidence. Knowing from where and when multimedia content originated is critical for establishing the chain of custody and guaranteeing the integrity of digital evidence. This is especially crucial when dealing with cases of child exploitation or forced labor, where timely preservation of evidence can make or break a case. 

\section{Conclusion and Future Directions}
\label{sec:conclusion}
This systematic review paper unveils the intricate tapestry of knowledge and innovation at the nexus of computer vision techniques, multimedia geolocation, and human trafficking combat. The synthesis of a diverse body of literature across these interconnected fields reveals the challenges, advancements, and prospects within this multifaceted realm. While substantial strides have been made in the development of advanced algorithms, pipelines, and tools to assist in human trafficking investigations and prosecutions, and indeed, more broadly in other lawful investigations, varied challenges persist.

Undoubtedly, multimedia geolocation estimation remains an excellent, yet difficult high-level computer vision problem, especially when all efforts at geolocation fail using explicitly structured available data such as metadata. One of the major drawbacks to effective geolocation in human trafficking investigation is the broad scope of this complex and pervasive crime. Due to the fact that both multimedia evidence surfacing and trafficking scope are on a global scale, the search space for possible location-matching reduces the accuracy of these techniques. One tactic for mitigating the problem is the categorization of problem space into indoor and outdoor scenes, thereby leading to a more fine-tuned vision-based solution. With such a sub-defined problem space, researchers and practitioners can leverage relevant cues such as GPS and geotagged data, landmarks, skyline, overhead and satellite imagery, and the sun azimuth angle for outdoor scene geolocation while cues such as latent inherent signals and frequencies extraction, and regional distinct decor recognition can be used to effective geolocation of indoor scene.

While multimedia data could entail a wide variety of data formats such as image, video, text, and audio, the fusion of these to yield multimodal data makes geolocation a more approachable problem for computer vision techniques to extract actionable intelligence. Geolocating with only one data format such as metadata analysis or image or video content analysis at a time reduces the contextual information needed for precise location estimation, where digital media is increasingly used by traffickers for recruitment, coercion, and communication.

One of the most significant revelations of this review is the transformative role played by CNNs and deep learning techniques in advancing the state-of-the-art in multimedia geolocation. The application of deep learning models, with their ability to automatically learn discriminative features from vast datasets, has demonstrated remarkable promise in enhancing the accuracy and robustness of geolocation algorithms. Within the purview of deep learning, the adoption of classification and retrieval techniques has emerged as a viable route. The ability of CNNs to not only classify multimedia content, but also to find semantically related examples from enormous databases represents a paradigm shift in how geolocation is tackled. By using the power of feature extraction and semantic comprehension, these strategies provide a more holistic and context-aware approach to geolocation.

As traffickers increasingly use digital platforms to conduct their illegal activities, the capacity to extract, authenticate, and analyze multimedia evidence becomes critical in constructing strong cases against criminals. The combination of computer vision techniques and digital forensics technologies has proved the ability to uncover concealed evidence, trace digital traces, and provide key insights into people traffickers' operations. Furthermore, the incorporation of deep learning approaches capable of handling data variations indicates a promising avenue in addressing variational factors such as aging and changes in facial appearance due to stress, trauma, or the passage of time, which make it more difficult to identify and find victims in certain circumstances over time. With GAN allowing for the synthesis and prediction of victims' appearance and changes over time, analyses such as kinship will broaden research and innovation in trafficking identification and combat.

To solve these problems and fully realize the potential of computer vision in human trafficking investigations, interdisciplinary collaboration, and ongoing research efforts are required. The knowledge synthesis given in this study provides a solid platform for future research and innovation in the persistent goal of eradicating human trafficking and ensuring justice for victims. The promise of more accurate, efficient, and context-aware geolocation solutions has the potential to significantly improve the fight against human trafficking, urging continued research, collaboration, and innovation in the relentless pursuit of justice and the complete extinction of human trafficking.

\bibliographystyle{model6-num-names}
\bibliography{bibfile}

\newpage
\onecolumn
\appendix

\section{Classification of the Systematically Reviewed Literature (Electronic Supplement)}

\begin{table}[!h]
\caption{Listing and Classification of Included Literature}
\label{tab:legend}
\centering
\begin{tabular}{|>{\scriptsize}c|>{\scriptsize}l|}

\hline
\multicolumn{2}{|c|}{\footnotesize\textbf{Legend}} \\ \hline
\faCircle & High Relevancy/Applicability  \\ \hline
\rotatebox[origin=c]{90}{\faAdjust} & Medium Relevancy/Applicability \\ \hline
\faCircle[regular] & Low Relevancy/Applicability \\ \hline
- & No Relevancy/Applicability or None \\ \hline
\faBinoculars  & Retrieval \\ \hline
\cincludegraphics[scale=0.015]{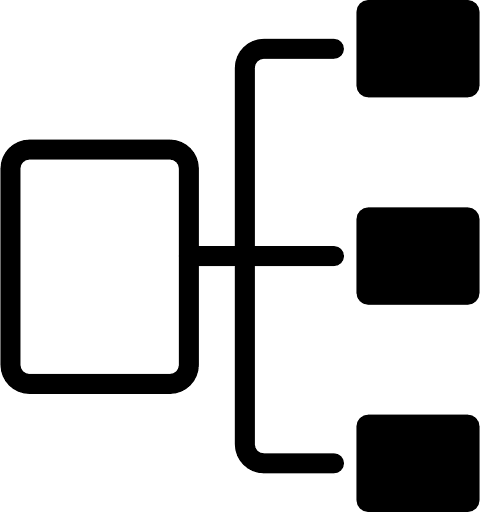}  & Classification \\ \hline
\cincludegraphics[scale=0.015]{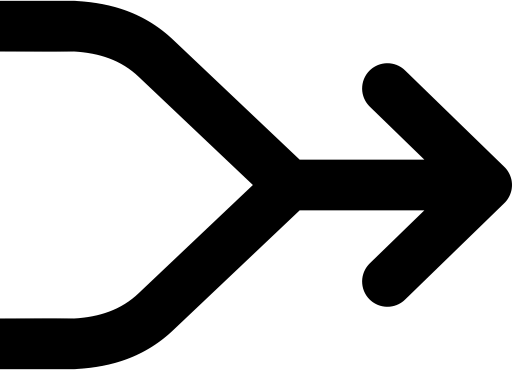}  & Combined \\ \hline
\faImage & Image \\ \hline
\faVideo & Video \\ \hline
\faMicrophone & Audio \\ \hline
\cincludegraphics[scale=0.012]{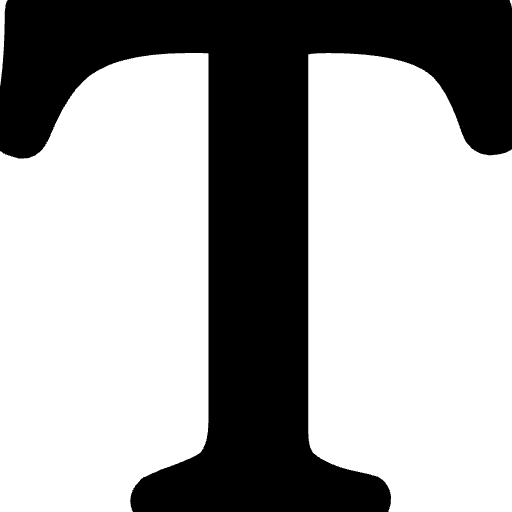} & Text \\ \hline
\cincludegraphics[scale=0.015]{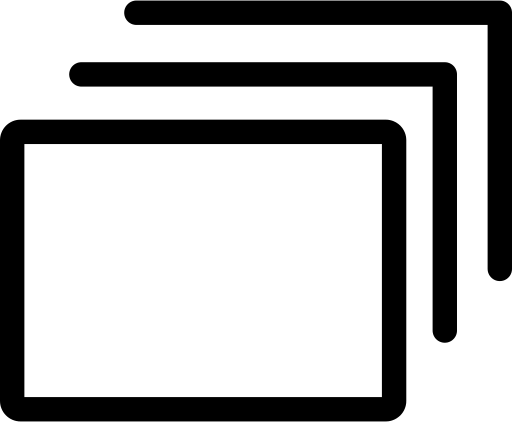}  & Multimodal \\ \hline

\cincludegraphics[scale=0.008]{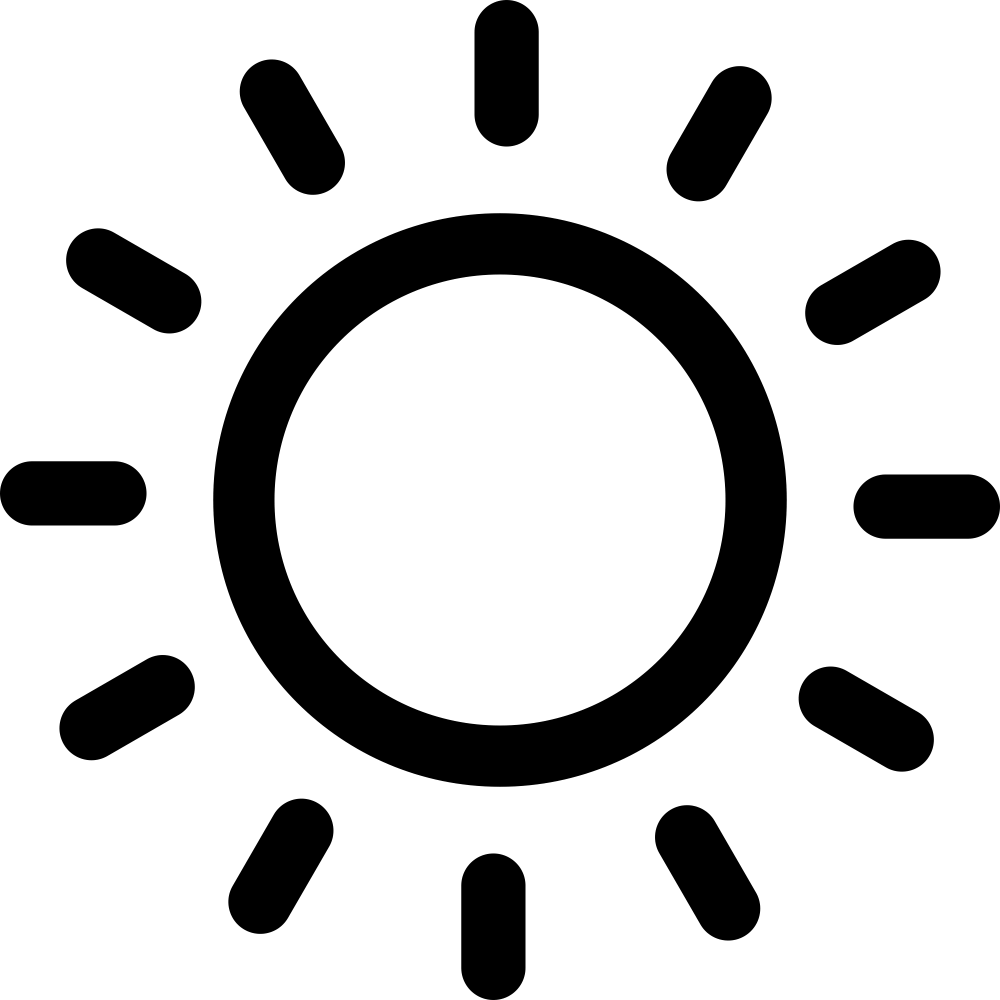} & Outdoor \\ \hline
\cincludegraphics[scale=0.02]{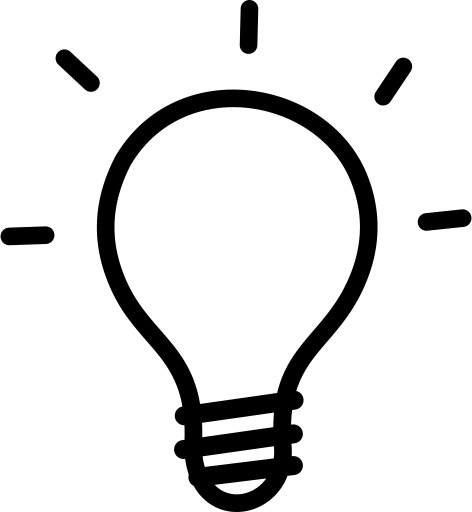} & Indoor \\ \hline

\end{tabular}
\end{table}


\small
\begin{longtable*}{|>{\footnotesize}p{5.7cm}|*{8}{>{\centering\arraybackslash\scriptsize}p{0.3cm}|}>{\centering\arraybackslash\scriptsize}p{0.6cm}|}
\hline
\textbf{Reference} & \rotatebox{90}{\textbf{Computer Vision}} & \rotatebox{90}{\textbf{Computer Vision Geolocation Technique~}} &  \rotatebox{90}{\textbf{Multimedia Geolocation}} & \rotatebox{90}{\textbf{Multimedia Data Type}} & \rotatebox{90}{\textbf{Digital Investigation}} & \rotatebox{90}{\textbf{Law Enforcement}} & \rotatebox{90}{\textbf{Human Trafficking}}  & \rotatebox{90}{\textbf{Hotel Room Identification}} &\rotatebox{90}{\textbf{Indoor/Outdoor}}   
\\ 
\hline
\endfirsthead

\multicolumn{10}{l}{\small \textsf{Table \ref{tab:legend}: Listing and Classification of Included Literature, continued}}\\[1ex]
\hline
\textbf{Reference} & \rotatebox{90}{\textbf{Computer Vision}} & \rotatebox{90}{\textbf{Computer Vision Geolocation Technique~}} &  \rotatebox{90}{\textbf{Multimedia Geolocation}} & \rotatebox{90}{\textbf{Multimedia Data Type}} & \rotatebox{90}{\textbf{Digital Investigation}} & \rotatebox{90}{\textbf{Law Enforcement}} & \rotatebox{90}{\textbf{Human Trafficking}}  & \rotatebox{90}{\textbf{Hotel Room Identification}} &\rotatebox{90}{\textbf{Indoor/Outdoor}}  
\\ 
\hline
\endhead
\hline
\endfoot




\citet{Glistrup_Urban-Image_2022}          &       \faCircle       &       \rotatebox[origin=c]{90}{\faAdjust}        &       \faCircle[regular]       &      -        &       -       &    -     & \cmark               &            \cincludegraphics[scale=0.015]{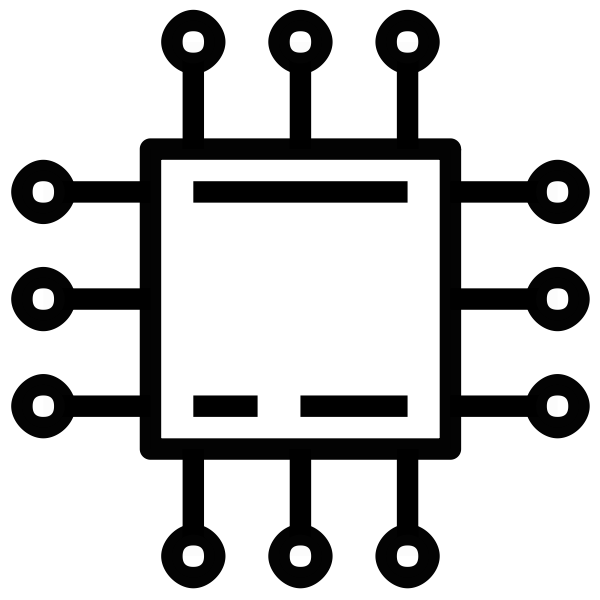}          &         \cincludegraphics[scale=0.008]{images/sun.png}\cincludegraphics[scale=0.02]{images/lightbulb.png}           \\ \hline

\citet{Sethi_Large-Scale_2013}       &       \rotatebox[origin=c]{90}{\faAdjust}       &       -       &       \faCircle[regular]       &       \cincludegraphics[scale=0.015]{images/multimedia.png}       &       \faCircle[regular]       &       \faCircle[regular]       &       \faCircle       &       -       &       -           \\ \hline

\citet{Sethi_Large-Scale_2013}       &       \rotatebox[origin=c]{90}{\faAdjust}       &       -       &       \faCircle[regular]       &       \cincludegraphics[scale=0.015]{images/multimedia.png}       &       \faCircle[regular]       &       \faCircle[regular]       &       \faCircle       &       -       &       -           \\ \hline

\citet{Glistrup_Urban-Image_2022}       &       \faCircle       &       \faBinoculars       &       \faCircle       &       \cincludegraphics[scale=0.015]{images/multimedia.png}       &       \rotatebox[origin=c]{90}{\faAdjust}       &       \rotatebox[origin=c]{90}{\faAdjust}       &       \faCircle[regular]       &       \rotatebox[origin=c]{90}{\faAdjust}       &       \cincludegraphics[scale=0.008]{images/sun.png}           \\ \hline

\citet{Stylianou_Indexing-open_2015}       &       \faCircle       &       \faBinoculars       &       \faCircle       &       \faImage       &       \faCircle       &       \rotatebox[origin=c]{90}{\faAdjust}       &       \faCircle       &       \faCircle       &       \cincludegraphics[scale=0.02]{images/lightbulb.png}           \\ \hline

\citet{Stylianou_TraffickCam_2017}       &       \faCircle       &       \cincludegraphics[scale=0.015]{images/classification-icon.png}       &       \faCircle       &       \faImage       &       \faCircle       &       \faCircle       &       \faCircle       &       \faCircle       &       \cincludegraphics[scale=0.02]{images/lightbulb.png}           \\ \hline

\citet{Yuan_geolocation_2022}       &       \faCircle       &       \cincludegraphics[scale=0.015]{images/classification-icon.png}       &       \faCircle       &       \faImage       &       \faCircle       &       \rotatebox[origin=c]{90}{\faAdjust}       &       \faCircle       &       \faCircle       &       \cincludegraphics[scale=0.02]{images/lightbulb.png}           \\ \hline

\citet{aronson_computer_2018}       &       \faCircle       &       -       &       \faCircle       &       \faVideo       &       \faCircle       &       \faCircle       &       \faCircle[regular]       &       -       &       \cincludegraphics[scale=0.008]{images/sun.png}           \\ \hline

\citet{Salem_Learning_2020}       &       \faCircle       &       -       &       \faCircle       &       \faImage       &       \rotatebox[origin=c]{90}{\faAdjust}       &       \rotatebox[origin=c]{90}{\faAdjust}       &       \faCircle[regular]       &       -       &       \cincludegraphics[scale=0.008]{images/sun.png}           \\ \hline

\citet{Ivanov_Comparative-Study_2013}       &       \faCircle       &       -       &       \faCircle       &       \cincludegraphics[scale=0.015]{images/multimedia.png}       &       \rotatebox[origin=c]{90}{\faAdjust}       &       \faCircle[regular]       &       -       &       -       &       \cincludegraphics[scale=0.008]{images/sun.png}           \\ \hline

\citet{bessinger2019generative}       &       \faCircle       &       -       &       \rotatebox[origin=c]{90}{\faAdjust}       &       \faImage       &       \rotatebox[origin=c]{90}{\faAdjust}       &       \faCircle[regular]       &       -       &       -       &       \cincludegraphics[scale=0.008]{images/sun.png}           \\ \hline

\citet{li_validating_2018}       &       \faCircle[regular]       &       -       &       \rotatebox[origin=c]{90}{\faAdjust}       &       \faImage       &       \faCircle       &       -       &       -       &       -       &       \cincludegraphics[scale=0.008]{images/sun.png}           \\ \hline

\citet{Kakar_Authenticating_2012}       &       \faCircle[regular]       &       -       &       \rotatebox[origin=c]{90}{\faAdjust}       &       \faImage       &       \faCircle       &       -       &       -       &       -       &       \cincludegraphics[scale=0.008]{images/sun.png}           \\ \hline

\citet{salem_timestamp_2022}       &       \faCircle       &       -       &       \faCircle       &       \faImage       &       \faCircle       &       \faCircle[regular]       &       -       &       -       &       \cincludegraphics[scale=0.008]{images/sun.png}           \\ \hline

\citet{Mookdarsanit_LocationEO_2017}       &       \faCircle[regular]       &       -       &       \faCircle       &       \cincludegraphics[scale=0.015]{images/multimedia.png}       &       \rotatebox[origin=c]{90}{\faAdjust}       &       \faCircle[regular]       &       -       &       -       &       \cincludegraphics[scale=0.008]{images/sun.png}           \\ \hline

\citet{Friedland_Multimodal_2010}       &       \faCircle       &       \cincludegraphics[scale=0.015]{images/combine-merge-arrow-icon.png}       &       \faCircle       &       \cincludegraphics[scale=0.015]{images/multimedia.png}       &       \faCircle       &       \faCircle       &       -       &       -       &       \cincludegraphics[scale=0.008]{images/sun.png}           \\ \hline

\citet{Padilha_Content-Aware_2022}       &       \faCircle       &       -       &       \faCircle       &       \cincludegraphics[scale=0.012]{images/t.png}       &       \faCircle       &       \faCircle[regular]       &       -       &       -       &       \cincludegraphics[scale=0.008]{images/sun.png}           \\ \hline

\citet{Ivasic-Kos_Application_2022}       &       \rotatebox[origin=c]{90}{\faAdjust}       &       \cincludegraphics[scale=0.015]{images/combine-merge-arrow-icon.png}       &       \faCircle[regular]       &       \cincludegraphics[scale=0.015]{images/multimedia.png}       &       \faCircle       &       \faCircle       &       \faCircle[regular]       &       -       &       \cincludegraphics[scale=0.008]{images/sun.png}           \\ \hline

\citet{moreira_image_2018}       &       \faCircle[regular]       &       -       &       \faCircle[regular]       &       \cincludegraphics[scale=0.012]{images/t.png}       &       \faCircle       &       \faCircle[regular]       &       -       &       -       &       \cincludegraphics[scale=0.008]{images/sun.png}           \\ \hline

\citet{Zhang_Landmark_2018}       &       \faCircle       &       \cincludegraphics[scale=0.015]{images/combine-merge-arrow-icon.png}       &       \faCircle       &       \cincludegraphics[scale=0.015]{images/multimedia.png}       &       \faCircle[regular]       &       -       &       -       &       -       &       \cincludegraphics[scale=0.008]{images/sun.png}           \\ \hline

\citet{Li_AreYouLying_2017}       &       \faCircle[regular]       &       -       &       \faCircle       &       \faImage       &       \faCircle       &       \faCircle       &       -       &       -       &       \cincludegraphics[scale=0.008]{images/sun.png}           \\ \hline

\citet{Nam_Revisiting_2017}       &       \faCircle       &       \cincludegraphics[scale=0.015]{images/combine-merge-arrow-icon.png}       &       \faCircle       &       \cincludegraphics[scale=0.015]{images/multimedia.png}       &       \rotatebox[origin=c]{90}{\faAdjust}       &       \rotatebox[origin=c]{90}{\faAdjust}       &       \faCircle[regular]       &       \faCircle[regular]       &       \cincludegraphics[scale=0.008]{images/sun.png}           \\ \hline

\citet{Choi_Invisible_2022}       &       \faCircle[regular]       &       -       &       \faCircle       &       \faImage       &       \faCircle       &       \faCircle[regular]       &       -       &       -       &       \cincludegraphics[scale=0.008]{images/sun.png}\cincludegraphics[scale=0.02]{images/lightbulb.png}           \\ \hline

\citet{Zhu_Large-Scale_2022}       &       \faCircle       &       \cincludegraphics[scale=0.015]{images/combine-merge-arrow-icon.png}       &       \faCircle       &       \faImage       &       \faCircle       &       \faCircle[regular]       &       -       &       -       &       \cincludegraphics[scale=0.008]{images/sun.png}           \\ \hline

\citet{Ma_Multi-source_2018}       &       \faCircle       &       \faBinoculars       &       \faCircle       &       \faImage       &       \faCircle       &       \faCircle[regular]       &       -       &       -       &       \cincludegraphics[scale=0.008]{images/sun.png}           \\ \hline

\citet{Piras_Information_2017}       &       \faCircle       &       \faBinoculars       &       \faCircle[regular]       &       \cincludegraphics[scale=0.015]{images/multimedia.png}       &       \faCircle       &       -       &       -       &       -       &       -           \\ \hline

\citet{Yokota_Visual-based_2020}       &       \faCircle       &       \cincludegraphics[scale=0.015]{images/combine-merge-arrow-icon.png}       &       \faCircle       &       \faImage       &       \faCircle       &       \faCircle       &       -       &       -       &       \cincludegraphics[scale=0.008]{images/sun.png}           \\ \hline

\citet{Al-Kuwari_Probabilistic-2010}       &       \faCircle[regular]       &       -       &       \rotatebox[origin=c]{90}{\faAdjust}       &       \cincludegraphics[scale=0.015]{images/multimedia.png}       &       \faCircle       &       \faCircle       &       -       &       -       &       \cincludegraphics[scale=0.008]{images/sun.png}           \\ \hline

\citet{Choi_Multimodal_2015}       &       \faCircle       &       \cincludegraphics[scale=0.015]{images/combine-merge-arrow-icon.png}       &       \faCircle       &       \cincludegraphics[scale=0.015]{images/multimedia.png}       &       \faCircle       &       \rotatebox[origin=c]{90}{\faAdjust}       &       -       &       -       &       \cincludegraphics[scale=0.008]{images/sun.png}\cincludegraphics[scale=0.02]{images/lightbulb.png}           \\ \hline

\citet{Neal_Global_2019}       &       \faCircle       &       \cincludegraphics[scale=0.015]{images/classification-icon.png}       &       \faCircle       &       \faImage       &       \faCircle       &       \faCircle       &       \faCircle[regular]       &       -       &       -           \\ \hline

\citet{Claudia_Live-forensics_2016}       &       \faCircle[regular]       &       \cincludegraphics[scale=0.015]{images/classification-icon.png}       &       \rotatebox[origin=c]{90}{\faAdjust}       &       \faImage       &       \faCircle       &       \rotatebox[origin=c]{90}{\faAdjust}       &       \faCircle[regular]       &       -       &       \cincludegraphics[scale=0.008]{images/sun.png}           \\ \hline

\citet{lallie_geotract_2013}       &       \faCircle[regular]       &       -       &       \faCircle       &       \cincludegraphics[scale=0.015]{images/multimedia.png}       &       \faCircle       &       \faCircle       &       \faCircle       &       -       &       \cincludegraphics[scale=0.008]{images/sun.png}\cincludegraphics[scale=0.02]{images/lightbulb.png}           \\ \hline

\citet{Lefèvre_Toward_2017}       &       \faCircle       &       \cincludegraphics[scale=0.015]{images/classification-icon.png}       &       \faCircle       &       \faImage       &       \rotatebox[origin=c]{90}{\faAdjust}       &       \faCircle[regular]       &       -       &       -       &       \cincludegraphics[scale=0.008]{images/sun.png}           \\ \hline

\citet{Anastasia_News_2020}       &       \faCircle[regular]       &       -       &       \faCircle[regular]       &       \faImage       &       \faCircle       &       \rotatebox[origin=c]{90}{\faAdjust}       &       -       &       -       &       -           \\ \hline

\citet{Roca_Novel_2016}       &       -       &       -       &       \faCircle[regular]       &       \faImage       &       \faCircle[regular]       &       \faCircle[regular]       &       -       &       -       &       \cincludegraphics[scale=0.008]{images/sun.png}           \\ \hline

\citet{Worring_Multimedia_2016}       &       \faCircle[regular]       &       -       &       \rotatebox[origin=c]{90}{\faAdjust}       &       \faImage       &       \faCircle[regular]       &       \faCircle[regular]       &       -       &       -       &       \cincludegraphics[scale=0.008]{images/sun.png}           \\ \hline

\citet{Ilčev_Global_2019}       &       \rotatebox[origin=c]{90}{\faAdjust}       &       -       &       \faCircle       &       \faImage       &       \rotatebox[origin=c]{90}{\faAdjust}       &       \faCircle[regular]       &       -       &       -       &       \cincludegraphics[scale=0.008]{images/sun.png}           \\ \hline

\citet{Liangzhi_A-deep-learning-semantic_2021}       &       \faCircle       &       -       &       \faCircle[regular]       &       \faImage       &       \faCircle[regular]       &       -       &       -       &       -       &       \cincludegraphics[scale=0.008]{images/sun.png}           \\ \hline

\citet{Jiang_Hierarchy-Dependent_2019}       &       \faCircle       &       \cincludegraphics[scale=0.015]{images/classification-icon.png}       &       \faCircle       &       \cincludegraphics[scale=0.015]{images/multimedia.png}       &       \rotatebox[origin=c]{90}{\faAdjust}       &       \faCircle[regular]       &       -       &       -       &       \cincludegraphics[scale=0.008]{images/sun.png}\cincludegraphics[scale=0.02]{images/lightbulb.png}           \\ \hline

\citet{Zerdoumi_A-new-spatial_2022}       &       \rotatebox[origin=c]{90}{\faAdjust}       &       -       &       \faCircle       &       \cincludegraphics[scale=0.015]{images/multimedia.png}       &       \rotatebox[origin=c]{90}{\faAdjust}       &       \faCircle[regular]       &       -       &       -       &       \cincludegraphics[scale=0.008]{images/sun.png}           \\ \hline

\citet{Keerthi_RIPA-2018}       &       \faCircle       &       \cincludegraphics[scale=0.015]{images/classification-icon.png}       &       \faCircle[regular]       &       \cincludegraphics[scale=0.015]{images/multimedia.png}       &       \rotatebox[origin=c]{90}{\faAdjust}       &       \rotatebox[origin=c]{90}{\faAdjust}       &       -       &       -       &       \cincludegraphics[scale=0.008]{images/sun.png}           \\ \hline

\citet{Kraus_Toward_2020}       &       \faCircle       &       -       &       \faCircle[regular]       &       \cincludegraphics[scale=0.015]{images/multimedia.png}       &       \rotatebox[origin=c]{90}{\faAdjust}       &       \faCircle       &       -       &       -       &       \cincludegraphics[scale=0.008]{images/sun.png}\cincludegraphics[scale=0.02]{images/lightbulb.png}           \\ \hline

\citet{Nilwong_DeepLearning_2019}       &       \faCircle       &       \cincludegraphics[scale=0.015]{images/classification-icon.png}       &       \faCircle       &       \faImage       &       \rotatebox[origin=c]{90}{\faAdjust}       &       -       &       -       &       -       &       \cincludegraphics[scale=0.008]{images/sun.png}           \\ \hline

\citet{workman_revisiting_2022}       &       \faCircle       &       -       &       \faCircle       &       \cincludegraphics[scale=0.015]{images/multimedia.png}       &       \rotatebox[origin=c]{90}{\faAdjust}       &       \rotatebox[origin=c]{90}{\faAdjust}       &       -       &       -       &       \cincludegraphics[scale=0.008]{images/sun.png}           \\ \hline

\citet{kuznetsov2015new}       &       \rotatebox[origin=c]{90}{\faAdjust}       &       -       &       \rotatebox[origin=c]{90}{\faAdjust}       &       \cincludegraphics[scale=0.015]{images/multimedia.png}       &       \rotatebox[origin=c]{90}{\faAdjust}       &       \rotatebox[origin=c]{90}{\faAdjust}       &       -       &       -       &       \cincludegraphics[scale=0.008]{images/sun.png}           \\ \hline

\citet{Smelyakov_Search-by-Image_2018}       &       \faCircle       &       \cincludegraphics[scale=0.015]{images/combine-merge-arrow-icon.png}       &       -       &       \faImage       &       -       &       -       &       -       &       -       &       \cincludegraphics[scale=0.008]{images/sun.png}\cincludegraphics[scale=0.02]{images/lightbulb.png}           \\ \hline

\citet{iqbal_machine_2020}       &       \faCircle       &       \cincludegraphics[scale=0.015]{images/combine-merge-arrow-icon.png}       &       \faCircle[regular]       &       \cincludegraphics[scale=0.015]{images/multimedia.png}       &       \faCircle       &       \rotatebox[origin=c]{90}{\faAdjust}       &       -       &       -       &       -           \\ \hline

\citet{Makantasis_Semi-supervised_2015}       &       \faCircle       &       \cincludegraphics[scale=0.015]{images/classification-icon.png}       &       \rotatebox[origin=c]{90}{\faAdjust}       &       \faImage       &       \faCircle       &       \rotatebox[origin=c]{90}{\faAdjust}       &       -       &       -       &       \cincludegraphics[scale=0.008]{images/sun.png}           \\ \hline

\citet{Aspandi_Heatmap-Guided_2019}       &       \faCircle       &       \cincludegraphics[scale=0.015]{images/classification-icon.png}       &       \faCircle[regular]       &       \faImage       &       \rotatebox[origin=c]{90}{\faAdjust}       &       \rotatebox[origin=c]{90}{\faAdjust}       &       \rotatebox[origin=c]{90}{\faAdjust}       &       -       &       \cincludegraphics[scale=0.008]{images/sun.png}           \\ \hline

\citet{deshmukh2022human}       &       \faCircle       &       -       &       \faCircle[regular]       &       \faImage       &       \faCircle[regular]       &       \rotatebox[origin=c]{90}{\faAdjust}       &       -       &       -       &       -           \\ \hline

\citet{Xiao_Tiny-object_2023}       &       \faCircle       &       -       &       \faCircle[regular]       &       \faImage       &       \rotatebox[origin=c]{90}{\faAdjust}       &       \rotatebox[origin=c]{90}{\faAdjust}       &       \rotatebox[origin=c]{90}{\faAdjust}       &       -       &       \cincludegraphics[scale=0.008]{images/sun.png}\cincludegraphics[scale=0.02]{images/lightbulb.png}           \\ \hline

\citet{Wu_Leveraging_2022}       &       \faCircle       &       -       &       -       &       \faImage       &       \faCircle[regular]       &       -       &       -       &       -       &       -           \\ \hline

\citet{Ammatmanee_Transfer-learning_2021}       &       \faCircle       &       \cincludegraphics[scale=0.015]{images/classification-icon.png}       &       \faCircle       &       \faImage       &       \rotatebox[origin=c]{90}{\faAdjust}       &       \rotatebox[origin=c]{90}{\faAdjust}       &       \rotatebox[origin=c]{90}{\faAdjust}       &       \rotatebox[origin=c]{90}{\faAdjust}       &       \cincludegraphics[scale=0.02]{images/lightbulb.png}           \\ \hline

\citet{Shao_HRSiam_2021}       &       \faCircle[regular]       &       -       &       \rotatebox[origin=c]{90}{\faAdjust}       &       \faImage       &       \faCircle[regular]       &       \rotatebox[origin=c]{90}{\faAdjust}       &       \faCircle[regular]       &       -       &       \cincludegraphics[scale=0.008]{images/sun.png}           \\ \hline

\citet{Shao_Can-We-Track_2019}       &       \rotatebox[origin=c]{90}{\faAdjust}       &       -       &       \rotatebox[origin=c]{90}{\faAdjust}       &       \faVideo       &       \rotatebox[origin=c]{90}{\faAdjust}       &       \rotatebox[origin=c]{90}{\faAdjust}       &       \faCircle[regular]       &       -       &       \cincludegraphics[scale=0.008]{images/sun.png}           \\ \hline

\citet{Liu_Age-Invariant_2022}       &       \rotatebox[origin=c]{90}{\faAdjust}       &       -       &       \faCircle[regular]       &       \faImage       &       \rotatebox[origin=c]{90}{\faAdjust}       &       \rotatebox[origin=c]{90}{\faAdjust}       &       \rotatebox[origin=c]{90}{\faAdjust}       &       -       &       -           \\ \hline

\citet{Karnik_A-New-Deep-Model_2022}       &       \faCircle       &       -       &       \faCircle[regular]       &       \faImage       &       \rotatebox[origin=c]{90}{\faAdjust}       &       \rotatebox[origin=c]{90}{\faAdjust}       &       \rotatebox[origin=c]{90}{\faAdjust}       &       -       &       -           \\ \hline

\citet{DBLP:conf/iciot2/YangHPXL20}       &       \faCircle       &       -       &       \faCircle[regular]       &       \faImage       &       \rotatebox[origin=c]{90}{\faAdjust}       &       \faCircle[regular]       &       \faCircle[regular]       &       -       &       -           \\ \hline

\citet{dimas2022_Operations-research}       &       -       &       -       &       \faCircle[regular]       &       -       &       \rotatebox[origin=c]{90}{\faAdjust}       &       \faCircle       &       \faCircle       &       \faCircle[regular]       &       -           \\ \hline

\citet{Hundman_Always-Lurking2018}       &       \rotatebox[origin=c]{90}{\faAdjust}       &       \cincludegraphics[scale=0.015]{images/classification-icon.png}       &       \faCircle[regular]       &       \cincludegraphics[scale=0.015]{images/multimedia.png}       &       \faCircle       &       \rotatebox[origin=c]{90}{\faAdjust}       &       \faCircle       &       -       &       \cincludegraphics[scale=0.008]{images/sun.png}\cincludegraphics[scale=0.02]{images/lightbulb.png}           \\ \hline

\citet{Hemadri_ChildrEN-SafEty_2022}       &       \faCircle       &       \faBinoculars       &       \rotatebox[origin=c]{90}{\faAdjust}       &       \faImage       &       \faCircle       &       \rotatebox[origin=c]{90}{\faAdjust}       &       \faCircle       &       -       &       -           \\ \hline

\citet{Raets_Trafficking_2021}       &       \faCircle[regular]       &       -       &       \rotatebox[origin=c]{90}{\faAdjust}       &       -       &       \faCircle       &       \faCircle       &       \faCircle       &       -       &       -           \\ \hline

\citet{Anda_Improving-Borderline_2019}       &       \faCircle       &       -       &       \faCircle[regular]       &       \faImage       &       \rotatebox[origin=c]{90}{\faAdjust}       &       \faCircle       &       \faCircle       &       -       &       -           \\ \hline

\citet{Serraoui_Knowledge-based_2022}       &       \faCircle       &       -       &       \faCircle[regular]       &       \faImage       &       \rotatebox[origin=c]{90}{\faAdjust}       &       \rotatebox[origin=c]{90}{\faAdjust}       &       \rotatebox[origin=c]{90}{\faAdjust}       &       -       &       -           \\ \hline

\citet{Wang_Cross-Generation_2019}       &       \faCircle       &       -       &       \faCircle[regular]       &       \faImage       &       \rotatebox[origin=c]{90}{\faAdjust}       &       \rotatebox[origin=c]{90}{\faAdjust}       &       \rotatebox[origin=c]{90}{\faAdjust}       &       -       &       -           \\ \hline

\citet{Chandaliya_Conditional_2019}       &       \faCircle       &       -       &       \faCircle[regular]       &       \faImage       &       \rotatebox[origin=c]{90}{\faAdjust}       &       \rotatebox[origin=c]{90}{\faAdjust}       &       \rotatebox[origin=c]{90}{\faAdjust}       &       -       &       -           \\ \hline

\citet{Upadhayay2021CombatingHT}       &       \faCircle[regular]       &       -       &       \faCircle[regular]       &       \cincludegraphics[scale=0.015]{images/multimedia.png}       &       \rotatebox[origin=c]{90}{\faAdjust}       &       \rotatebox[origin=c]{90}{\faAdjust}       &       \faCircle       &       -       &       -           \\ \hline

\citet{tong_combating_2017}       &       \faCircle       &       \cincludegraphics[scale=0.015]{images/classification-icon.png}       &       \faCircle[regular]       &       \cincludegraphics[scale=0.015]{images/multimedia.png}       &       \rotatebox[origin=c]{90}{\faAdjust}       &       \rotatebox[origin=c]{90}{\faAdjust}       &       \faCircle       &       -       &       -           \\ \hline

\citet{Goyal_Family_2022}       &       \faCircle       &       \cincludegraphics[scale=0.015]{images/classification-icon.png}       &       \faCircle[regular]       &       \faImage       &       \faCircle       &       \rotatebox[origin=c]{90}{\faAdjust}       &       \rotatebox[origin=c]{90}{\faAdjust}       &       -       &       -           \\ \hline

\citet{Goh_Findind_2021}       &       -       &       -       &       -       &       \faImage       &       \rotatebox[origin=c]{90}{\faAdjust}       &       \rotatebox[origin=c]{90}{\faAdjust}       &       \faCircle       &       -       &       -           \\ \hline

\citet{Mukherjee_A-Machine_2021}       &       \faCircle       &       \cincludegraphics[scale=0.015]{images/classification-icon.png}       &       \faCircle       &       \faImage       &       \faCircle       &       \rotatebox[origin=c]{90}{\faAdjust}       &       \faCircle       &       -       &       \cincludegraphics[scale=0.008]{images/sun.png}           \\ \hline

\citet{sierra2022twitter}       &       \faCircle[regular]       &       -       &       -       &       \cincludegraphics[scale=0.012]{images/t.png}       &       \rotatebox[origin=c]{90}{\faAdjust}       &       \faCircle[regular]       &       \faCircle       &       -       &       -           \\ \hline

\citet{Matasci_Deep_2021}       &       \faCircle       &       \cincludegraphics[scale=0.015]{images/classification-icon.png}       &       \rotatebox[origin=c]{90}{\faAdjust}       &       \faImage       &       \faCircle       &       \rotatebox[origin=c]{90}{\faAdjust}       &       \rotatebox[origin=c]{90}{\faAdjust}       &       -       &       \cincludegraphics[scale=0.008]{images/sun.png}           \\ \hline

\citet{abbas2022kinship}       &       \rotatebox[origin=c]{90}{\faAdjust}       &       \cincludegraphics[scale=0.015]{images/classification-icon.png}       &       \faCircle[regular]       &       \faImage       &       \rotatebox[origin=c]{90}{\faAdjust}       &       \rotatebox[origin=c]{90}{\faAdjust}       &       \rotatebox[origin=c]{90}{\faAdjust}       &       -       &       -           \\ \hline

\citet{Karpagam_A-novel-2022}       &       \faCircle       &       \cincludegraphics[scale=0.015]{images/classification-icon.png}       &       \rotatebox[origin=c]{90}{\faAdjust}       &       \faVideo       &       \rotatebox[origin=c]{90}{\faAdjust}       &       \rotatebox[origin=c]{90}{\faAdjust}       &       \faCircle       &       -       &       -           \\ \hline

\citet{Karnik_A-new_2022}       &       \faCircle       &       \cincludegraphics[scale=0.015]{images/classification-icon.png}       &       \faCircle[regular]       &       \faImage       &       \rotatebox[origin=c]{90}{\faAdjust}       &       \faCircle       &       \faCircle       &       -       &       -           \\ \hline

\citet{Haibin_Multi-scale_2021}       &       \faCircle       &       \cincludegraphics[scale=0.015]{images/classification-icon.png}       &       -       &       \faImage       &       \rotatebox[origin=c]{90}{\faAdjust}       &       \rotatebox[origin=c]{90}{\faAdjust}       &       \faCircle       &       -       &       -           \\ \hline

\citet{Kshetri_regulatory_2023}       &       -       &       -       &       -       &       -       &       \faCircle[regular]       &       \faCircle[regular]       &       \faCircle[regular]       &       -       &       -           \\ \hline

\citet{Koudelová_Modelling_2014}       &       \rotatebox[origin=c]{90}{\faAdjust}       &       -       &       \rotatebox[origin=c]{90}{\faAdjust}       &       \faImage       &       \rotatebox[origin=c]{90}{\faAdjust}       &       \rotatebox[origin=c]{90}{\faAdjust}       &       \rotatebox[origin=c]{90}{\faAdjust}       &       -       &       -           \\ \hline

\citet{Paul_Multi-modal_2020}       &       \faCircle       &       \cincludegraphics[scale=0.015]{images/classification-icon.png}       &       \faCircle[regular]       &       \faImage       &       \rotatebox[origin=c]{90}{\faAdjust}       &       \rotatebox[origin=c]{90}{\faAdjust}       &       \faCircle       &       -       &       -           \\ \hline

\citet{Sukumar_Open-research_2015}       &       \faCircle       &       -       &       \faCircle       &       \faImage       &       \rotatebox[origin=c]{90}{\faAdjust}       &       \rotatebox[origin=c]{90}{\faAdjust}       &       \rotatebox[origin=c]{90}{\faAdjust}       &       -       &       \cincludegraphics[scale=0.008]{images/sun.png}\cincludegraphics[scale=0.02]{images/lightbulb.png}           \\ \hline

\citet{Goyal_Patch-Based_2021}       &       \faCircle       &       -       &       \faCircle[regular]       &       \faImage       &       \faCircle[regular]       &       \faCircle[regular]       &       \rotatebox[origin=c]{90}{\faAdjust}       &       -       &       -           \\ \hline

\citet{Murdock_2016}       &       \faCircle[regular]       &       -       &       \rotatebox[origin=c]{90}{\faAdjust}       &       \cincludegraphics[scale=0.012]{images/t.png}       &       \rotatebox[origin=c]{90}{\faAdjust}       &       \rotatebox[origin=c]{90}{\faAdjust}       &       \faCircle       &       -       &       -           \\ \hline

\citet{kamath20212021}       &       \faCircle[regular]       &       \cincludegraphics[scale=0.015]{images/classification-icon.png}       &       \rotatebox[origin=c]{90}{\faAdjust}       &       \faImage       &       \rotatebox[origin=c]{90}{\faAdjust}       &       \faCircle[regular]       &       \faCircle       &       \faCircle       &       \cincludegraphics[scale=0.02]{images/lightbulb.png}           \\ \hline

\citet{Mattmann_Scalable_2017}       &       \faCircle[regular]       &       -       &       \faCircle[regular]       &       \faVideo       &       \faCircle       &       \rotatebox[origin=c]{90}{\faAdjust}       &       \faCircle       &       -       &       \cincludegraphics[scale=0.008]{images/sun.png}\cincludegraphics[scale=0.02]{images/lightbulb.png}           \\ \hline

\citet{Chen_Semi-coupled_2021}       &       \rotatebox[origin=c]{90}{\faAdjust}       &       -       &       \faCircle[regular]       &       \faImage       &       \rotatebox[origin=c]{90}{\faAdjust}       &       \faCircle[regular]       &       \faCircle[regular]       &       -       &       -           \\ \hline

\citet{Casino_Sok_2022}       &       -       &       -       &       -       &       -       &       \faCircle       &       \faCircle       &       \rotatebox[origin=c]{90}{\faAdjust}       &       -       &       -           \\ \hline

\citet{Vajiac_TrafficVis_2023}       &       \faCircle[regular]       &       -       &       \rotatebox[origin=c]{90}{\faAdjust}       &       \cincludegraphics[scale=0.015]{images/multimedia.png}       &       \faCircle       &       \faCircle       &       \faCircle       &       -       &       \cincludegraphics[scale=0.008]{images/sun.png}           \\ \hline

\citet{Ramchandani_Unmasking_2021}       &       \faCircle[regular]       &       -       &       \faCircle       &       \cincludegraphics[scale=0.012]{images/t.png}       &       \rotatebox[origin=c]{90}{\faAdjust}       &       \rotatebox[origin=c]{90}{\faAdjust}       &       \faCircle       &       -       &       -           \\ \hline

\citet{Robinson_Visual_2018}       &       \faCircle       &       \cincludegraphics[scale=0.015]{images/classification-icon.png}       &       \rotatebox[origin=c]{90}{\faAdjust}       &       \faImage       &       \rotatebox[origin=c]{90}{\faAdjust}       &       \rotatebox[origin=c]{90}{\faAdjust}       &       \rotatebox[origin=c]{90}{\faAdjust}       &       -       &       -           \\ \hline

\citet{Robinson_To-Recognize_2018}       &       \faCircle       &       \cincludegraphics[scale=0.015]{images/classification-icon.png}       &       \rotatebox[origin=c]{90}{\faAdjust}       &       \faImage       &       \rotatebox[origin=c]{90}{\faAdjust}       &       \rotatebox[origin=c]{90}{\faAdjust}       &       \rotatebox[origin=c]{90}{\faAdjust}       &       -       &       -           \\ \hline

\citet{Shao_Tracking_2019}       &       \rotatebox[origin=c]{90}{\faAdjust}       &       -       &       \rotatebox[origin=c]{90}{\faAdjust}       &       \faVideo       &       \rotatebox[origin=c]{90}{\faAdjust}       &       \faCircle[regular]       &       \faCircle[regular]       &       -       &       \cincludegraphics[scale=0.008]{images/sun.png}           \\ \hline

\citet{Yadav_A-Feature-Averaging_2019}       &       \rotatebox[origin=c]{90}{\faAdjust}       &       \cincludegraphics[scale=0.015]{images/classification-icon.png}       &       \faCircle[regular]       &       \faImage       &       \rotatebox[origin=c]{90}{\faAdjust}       &       \faCircle[regular]       &       \faCircle[regular]       &       -       &       -           \\ \hline

\citet{dasgupta2022audio}       &       \rotatebox[origin=c]{90}{\faAdjust}       &       \cincludegraphics[scale=0.015]{images/classification-icon.png}       &       \faCircle[regular]       &       \faMicrophone       &       \rotatebox[origin=c]{90}{\faAdjust}       &       \faCircle       &       \faCircle       &       -       &       \cincludegraphics[scale=0.008]{images/sun.png}           \\ \hline

\citet{Anda_Assessing_2020}       &       \rotatebox[origin=c]{90}{\faAdjust}       &       -       &       \faCircle[regular]       &       \faImage       &       \faCircle       &       \faCircle       &       \rotatebox[origin=c]{90}{\faAdjust}       &       -       &       -           \\ \hline

\citet{Chandaliya_child_2021}       &       \faCircle       &       -       &       \faCircle[regular]       &       \faImage       &       \rotatebox[origin=c]{90}{\faAdjust}       &       \rotatebox[origin=c]{90}{\faAdjust}       &       \rotatebox[origin=c]{90}{\faAdjust}       &       -       &       -           \\ \hline

\citet{salamatcitesranger}       &       \faCircle       &       \cincludegraphics[scale=0.015]{images/classification-icon.png}       &       \faCircle[regular]       &       \faImage       &       \rotatebox[origin=c]{90}{\faAdjust}       &       \rotatebox[origin=c]{90}{\faAdjust}       &       \faCircle       &       -       &       \cincludegraphics[scale=0.008]{images/sun.png}           \\ \hline

\citet{Vinavatani_AI_2022}       &       \faCircle[regular]       &       \cincludegraphics[scale=0.015]{images/classification-icon.png}       &       \rotatebox[origin=c]{90}{\faAdjust}       &       \faImage       &       \rotatebox[origin=c]{90}{\faAdjust}       &       \faCircle       &       \faCircle       &       -       &       -           \\ \hline

\citet{Li_Detection_2018}       &       \rotatebox[origin=c]{90}{\faAdjust}       &       -       &       \rotatebox[origin=c]{90}{\faAdjust}       &       \cincludegraphics[scale=0.012]{images/t.png}       &       \faCircle       &       \rotatebox[origin=c]{90}{\faAdjust}       &       \faCircle       &       -       &       -           \\ \hline

\citet{Granizo_detection_2020}       &       \faCircle       &       -       &       \faCircle[regular]       &       \cincludegraphics[scale=0.012]{images/t.png}       &       \rotatebox[origin=c]{90}{\faAdjust}       &       \rotatebox[origin=c]{90}{\faAdjust}       &       \rotatebox[origin=c]{90}{\faAdjust}       &       -       &       -           \\ \hline

\citet{Deeb-Swihart_Ethical_2022}       &       \rotatebox[origin=c]{90}{\faAdjust}       &       -       &       -       &       -       &       \rotatebox[origin=c]{90}{\faAdjust}       &       \faCircle       &       \faCircle       &       -       &       -           \\ \hline

\citet{Zhang_DeepDive_2017}       &       -       &       -       &       -       &       -       &       \faCircle[regular]       &       \faCircle[regular]       &       \rotatebox[origin=c]{90}{\faAdjust}       &       -       &       -           \\ \hline

\citet{Mohamadi_Human_2021}       &       \rotatebox[origin=c]{90}{\faAdjust}       &       \cincludegraphics[scale=0.015]{images/classification-icon.png}       &       -       &       \cincludegraphics[scale=0.012]{images/t.png}       &       \rotatebox[origin=c]{90}{\faAdjust}       &       \faCircle[regular]       &       \rotatebox[origin=c]{90}{\faAdjust}       &       -       &       -           \\ \hline

\citet{Swaminathan_Predict_2022}       &       \faCircle       &       \cincludegraphics[scale=0.015]{images/classification-icon.png}       &       -       &       \cincludegraphics[scale=0.012]{images/t.png}       &       \faCircle       &       \rotatebox[origin=c]{90}{\faAdjust}       &       \faCircle[regular]       &       -       &       -           \\ \hline

\citet{vyawahare2022synthetic}       &       \faCircle       &       -       &       -       &       \faImage       &       \faCircle[regular]       &       \faCircle[regular]       &       \faCircle[regular]       &       -       &       -           \\ \hline

\citet{Varastehpour_Vein_2019}       &       \rotatebox[origin=c]{90}{\faAdjust}       &       -       &       -       &       \faImage       &       \faCircle       &       \rotatebox[origin=c]{90}{\faAdjust}       &       \faCircle[regular]       &       -       &       -           \\ \hline

\citet{chhoriya2019automated}       &       \faCircle       &       -       &       \faCircle[regular]       &       \faImage       &       \faCircle       &       \faCircle[regular]       &       \faCircle[regular]       &       -       &       -           \\ \hline

\citet{Sukumar_Open_2015}       &       \faCircle[regular]       &       -       &       -       &       -       &       \rotatebox[origin=c]{90}{\faAdjust}       &       \faCircle[regular]       &       \faCircle[regular]       &       -       &       -           \\ \hline

\citet{Kejriwal_technology_2018}       &       -       &       -       &       \faCircle[regular]       &       -       &       \faCircle       &       \faCircle       &       \rotatebox[origin=c]{90}{\faAdjust}       &       -       &       -           \\ \hline

\citet{Parolin_3M-Transformers_2021}       &       \rotatebox[origin=c]{90}{\faAdjust}       &       -       &       \faCircle[regular]       &       -       &       \faCircle       &       \rotatebox[origin=c]{90}{\faAdjust}       &       \rotatebox[origin=c]{90}{\faAdjust}       &       -       &       -           \\ \hline

\citet{Zavrsnik_criminal_2020}       &       \rotatebox[origin=c]{90}{\faAdjust}       &       -       &       -       &       -       &       \rotatebox[origin=c]{90}{\faAdjust}       &       \faCircle       &       \rotatebox[origin=c]{90}{\faAdjust}       &       -       &       -           \\ \hline

\citet{Horan_cyber_2021}       &       \faCircle[regular]       &       -       &       -       &       -       &       \rotatebox[origin=c]{90}{\faAdjust}       &       \rotatebox[origin=c]{90}{\faAdjust}       &       \rotatebox[origin=c]{90}{\faAdjust}       &       -       &       -           \\ \hline

\citet{Kim-Kwang_Forensic-Visualization_2017}       &       -       &       -       &       -       &       -       &       \faCircle       &       -       &       \faCircle[regular]       &       -       &       -           \\ \hline

\citet{Alruwaili_CustodyBlock_2021}       &       -       &       -       &       -       &       -       &       \faCircle       &       \faCircle[regular]       &       -       &       -       &       -           \\ \hline

\citet{bohme2016media}       &       \faCircle[regular]       &       -       &       \faCircle[regular]       &       \faImage       &       \faCircle       &       \faCircle[regular]       &       \faCircle[regular]       &       -       &       -           \\ \hline

\citet{GALBRAITH2020301009}       &       \faCircle[regular]       &       -       &       \faCircle       &       \cincludegraphics[scale=0.015]{images/multimedia.png}       &       \faCircle       &       \rotatebox[origin=c]{90}{\faAdjust}       &       \rotatebox[origin=c]{90}{\faAdjust}       &       -       &       \cincludegraphics[scale=0.008]{images/sun.png}           \\ \hline

\citet{Srivastava_CamForensics_2017}       &       \faCircle[regular]       &       -       &       \faCircle[regular]       &       \faImage       &       \faCircle       &       \faCircle[regular]       &       -       &       -       &       \cincludegraphics[scale=0.008]{images/sun.png}           \\ \hline

\citet{Sharifzadeh_Vein-Pattern_2014}       &       \rotatebox[origin=c]{90}{\faAdjust}       &       -       &       -       &       \faImage       &       \faCircle       &       \rotatebox[origin=c]{90}{\faAdjust}       &       \faCircle[regular]       &       -       &       -           \\ \hline

\citet{Varastehpour_Vein-Pattern_2019}       &       \rotatebox[origin=c]{90}{\faAdjust}       &       -       &       -       &       \faImage       &       \faCircle       &       \rotatebox[origin=c]{90}{\faAdjust}       &       \faCircle[regular]       &       -       &       -           \\ \hline

\citet{Keivanmarz_Vein-Pattern_2020}       &       \faCircle       &       -       &       -       &       \faImage       &       \faCircle       &       \rotatebox[origin=c]{90}{\faAdjust}       &       \faCircle[regular]       &       -       &       -           \\ \hline

\citet{Robinson_The_2021}       &       \faCircle       &       -       &       \faCircle[regular]       &       \faImage       &       \rotatebox[origin=c]{90}{\faAdjust}       &       \rotatebox[origin=c]{90}{\faAdjust}       &       \rotatebox[origin=c]{90}{\faAdjust}       &       -       &       -           \\ \hline

\citet{Gowda_An-Approach_2017}       &       \faCircle       &       \cincludegraphics[scale=0.015]{images/combine-merge-arrow-icon.png}       &       \faCircle[regular]       &       \cincludegraphics[scale=0.015]{images/multimedia.png}       &       \faCircle       &       \rotatebox[origin=c]{90}{\faAdjust}       &       \faCircle       &       -       &       \cincludegraphics[scale=0.008]{images/sun.png}\cincludegraphics[scale=0.02]{images/lightbulb.png}           \\ \hline

\citet{Anda_Evaluating_2018}       &       \faCircle[regular]       &       -       &       \faCircle[regular]       &       \faImage       &       \faCircle       &       \rotatebox[origin=c]{90}{\faAdjust}       &       \faCircle[regular]       &       -       &       -           \\ \hline

\citet{Black_Evaluation_2020}       &       \rotatebox[origin=c]{90}{\faAdjust}       &       -       &       \rotatebox[origin=c]{90}{\faAdjust}       &       \cincludegraphics[scale=0.015]{images/multimedia.png}       &       \faCircle       &       \faCircle[regular]       &       \faCircle[regular]       &       -       &       \cincludegraphics[scale=0.008]{images/sun.png}\cincludegraphics[scale=0.02]{images/lightbulb.png}           \\ \hline

\citet{Stylianou_Hotels-50K_2019}       &       \faCircle       &       \cincludegraphics[scale=0.015]{images/combine-merge-arrow-icon.png}       &       \faCircle       &       \faImage       &       \faCircle       &       \rotatebox[origin=c]{90}{\faAdjust}       &       \faCircle       &       \faCircle       &       \cincludegraphics[scale=0.02]{images/lightbulb.png}           \\ \hline

\end{longtable*}







\end{document}